\documentclass{article}

\PassOptionsToPackage{numbers, compress}{natbib}

\usepackage[final]{neurips_2022}

\usepackage[utf8]{inputenc} 
\usepackage[T1]{fontenc}    
\usepackage{hyperref}       
\usepackage{url}            
\usepackage{booktabs}       
\usepackage{amsfonts}       
\usepackage{nicefrac}       
\usepackage{microtype}      
\usepackage{xcolor}         

\usepackage{graphicx}
\usepackage{caption}
\usepackage{subfigure}
\usepackage{wrapfig}
\usepackage{algorithm}
\usepackage{algorithmic}

\usepackage{tikz}
\usetikzlibrary{bayesnet}
\usetikzlibrary{arrows}
\usetikzlibrary{positioning}
\usetikzlibrary{arrows,automata} 
\usetikzlibrary{external}
\usetikzlibrary{arrows,positioning} 
\tikzset{
    >=stealth',
    punkt/.style={
          rectangle,
          rounded corners,
          draw=black, very thick,
          text width=6.5em,
          minimum height=2em,
          text centered},
    pil/.style={
          ->,
          thick,
          shorten <=2pt,
          shorten >=2pt,}
}

\usepackage{amsmath,amsfonts,bm}

\newcommand{\D}{\mathcal{D}}

\newcommand{\R}{\mathbb{R}}

\newcommand{\E}{\mathbb{E}}

\newcommand{\calX}{\mathcal{X}}
\newcommand{\calA}{\mathcal{A}}
\newcommand{\calY}{\mathcal{Y}}
\newcommand{\calL}{\mathcal{L}}

\newcommand{\calN}{\mathcal{N}}

\newcommand{\RN}[1]{%
  \textup{\uppercase\expandafter{\romannumeral#1}}%
}

\usepackage{amsmath}
\usepackage{amssymb}
\usepackage{mathtools}
\usepackage{amsthm}
\usepackage{enumitem}

\newtheorem{theorem}{Theorem}[section]
\newtheorem{corollary}{Corollary}[section]
\newtheorem{lemma}[theorem]{Lemma}
\newtheorem{definition}{Definition}[section]
\newtheorem{proposition}[theorem]{Proposition}

\title{On Learning Fairness and Accuracy on Multiple Subgroups}

\author{
 Changjian Shui$^{1,2,4,*}$\quad 
 Gezheng Xu$^{3,}$\thanks{Equal contribution}\quad 
 Qi Chen$^{4}$\quad
 Jiaqi Li$^{3}$\quad \\
 \textbf{Charles X. Ling}$^{3}$\quad
 \textbf{Tal Arbel}$^{1,2,5}$\quad
 \textbf{Boyu Wang}$^{3,\dagger}$\quad
 \textbf{Christian Gagn\'e}$^{2,4,5,}$ \thanks{Corresponding}\\
  $^{1}$Centre for Intelligent Machines, McGill University \quad\quad 
  $^{2}$Mila, Quebec AI Institute\\
  $^{3}$Department of Computer Science, University of Western Ontario \\
  $^{4}$Institute Intelligence and Data, Université Laval \quad\quad
  $^{5}$CIFAR AI Chair \\
}

\begin{document}
\maketitle
\begin{abstract}
  We propose an analysis in fair learning that preserves the utility of the data while reducing prediction disparities under the criteria of group sufficiency. We focus on the scenario where the data contains multiple or even many subgroups, each with \emph{limited number} of samples. As a result, we present a principled method for learning a fair predictor for all subgroups via formulating it as a bilevel objective. In the lower-level, the subgroup-specific predictors are learned through a small amount of data and the fair predictor. In the upper-level, the fair predictor is updated to be close to all subgroup specific predictors. We further prove that such a bilevel objective can effectively control the group sufficiency and generalization error. We evaluate the proposed framework on real-world datasets. Empirical evidence suggests the consistently improved fair predictions, as well as the comparable accuracy to the baselines.
\end{abstract}

\section{Introduction}
Machine learning has made rapid progress in sociotechnical systems
such as automatic resume screening, video surveillance, and credit scoring for loan applications. Simultaneously, it has been observed that learning algorithms exhibited biased predictions on the \emph{subgroups} of population \citep{barocas-hardt-narayanan,kearns2018preventing}. For example, the algorithm denies a loan application based on sensitive attributes 
such as gender, race, or disability, which has heightened public concerns. 

To this end, fair learning is recently highlighted to mitigate prediction disparities. The high-level idea is quite straightforward: adding fair constraints during the training  \citep{donini2018empirical}. As a result, fair learning principally gives rise to two desiderata. On the one hand, the fair predictor should be \emph{informative} to ensure  accurate predictions for the data. On the other hand, the predictor is required to guarantee fairness to avoid prediction disparities across subgroups. Therefore, it is crucial to understand the possibilities and then design provable approaches for achieving both \emph{informative} and \emph{fair} learning. 

Clearly, achieving both objectives depends on predefined fair notations. Consider demographic parity \cite{barocas-hardt-narayanan} as the fair criteria, which necessitates the independence between the predictor's output $f(X)$ and the sensitive attribute (or subgourp index) $A$. Thus, if the sensitive attribute $A$ and the ground-truth label $Y$ are highly correlated, it is impossible to learn a both fair and informative predictor. 

To avoid such intrinsic impossibilities, alternative fair notions have been developed. In this work, we focus on the criteria of \emph{group sufficiency} \citep{barocas-hardt-narayanan,chouldechova2017fair}, which ensures that the conditional expectation of ground-truth label  ($\E[Y|f(X),A]$) is identical across different subgroups, given the predictor's output. Notably, the risk of violating group sufficiency has arisen in a number of real-world scenarios. E.g., in medical artificial intelligence, the machine learning algorithm is used to assess the clinic risk, and guide decisions regarding initiating medical therapy. However, \cite{doi:10.1056/NEJMms2004740,doi:10.1126/science.aax2342} revealed a significant racial bias in such algorithms: when the algorithm predicts the same clinical risk score $f(X)$ for white and black patients, black patients are actually at a higher risk of severe illness: $\E[Y|f(X),A=\text{black}]\gg \E[Y|f(X),A=\text{white}]$. The deployed algorithms have resulted in more referrals of white patients to specialty healthcare services, resulting in both spending disparities and racial bias  \citep{doi:10.1056/NEJMms2004740}. 

In summary, this work aims to propose a novel principled framework for ensuring group sufficiency, as well as preserving an informative prediction with a small generalization error. In particular, we focus on one challenge scenario: the \emph{data includes multiple or even a large number of subgroups, some with only limited samples}, as often occurs in the real-world. For example, datasets for the self-driving car are collected from a wide range of geographical regions, each with a limited number of training samples \cite{FrankvanPraat2020}. How can we ensure group sufficiency as well as accurate predictions? Specifically, our contributions are summarized as follows:

\begin{wrapfigure}{r}{4.5cm}
    \centering





\includegraphics[width=40mm]{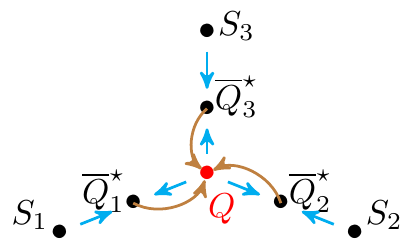}
\caption{Illustration of the proposed algorithm. Consider three subgroups $S_1, S_2, S_3$, e.g., datasets for three different races. The proposed algorithm is then formulated as a bilevel optimization to learn an informative and fair predictive-distribution $Q$. In the lower-level 
(\textcolor{cyan}{cyan}), we learn the subgroup specific predictive-distribution $\overline{Q}^{\star}_{a}$ from dataset $S_a$ (limited samples) and the prior $Q$. In the upper-level (\textcolor{brown}{brown}), $Q$ is then updated to be as close to all of the learned subgroup specific $\overline{Q}^{\star}_{a}$ as possible.}
    \label{fig:illus_opt}
\end{wrapfigure}

\textbf{Controlling group sufficiency}\quad
We adopted \emph{group sufficiency gap} to measure  fairness w.r.t. group sufficiency of a classifier $f$ (Sec.\ref{sec:def}), and then derive an \emph{upper bound} of the group sufficiency gap (Theorem \ref{thm:uff_upp}). Under proper assumptions, the upper bound is controlled by the discrepancy between the classifier $f$ and the subgroup Bayes predictors. Namely, minimizing the upper bound also encourages an informative classifier. 

\textbf{Algorithmic contribution}\quad
Motivated by the upper bound of the group sufficiency gap, we develop a principled algorithm. Concretely, we adopt a randomized algorithm that produces a predictive-distribution $Q$ over the classifier ($f \sim Q$) to learn informative and fair classification. 
We further formulate the problem as a bilevel optimization (Sec.~\ref{sec:bilevel}), as shown in Fig.\ref{fig:illus_opt}.
\textcolor{cyan}{(1)} In the lower-level, the subgroup specific dataset $S_a$ and the fair predictive-distribution $Q$ are used to learn the subgroup specific predictive-distribution $\overline{Q}^{\star}_a$, where $Q$ is regarded as an \emph{informative prior} for learning limited data within each subgroup. Theorem \ref{thm:pac_bayes} formally demonstrates that under proper assumptions, the lower-level loss can effectively control the generalization error. \textcolor{brown}{(2)}
In the upper-level, the fair predictive-distribution $Q$ is then updated to be close to all subgroup specific predictive-distributions, in order to minimize the upper bound of the group sufficiency gap. 

\textbf{Empirical justifications}~~The proposed algorithm is applicable to the general parametric and differentiable model, where we adopt the neural network in the implementation.
We evaluate the proposed algorithm on two real-world NLP datasets that have shown prediction disparities w.r.t. group sufficiency. Compared with baselines, the results indicate that group sufficiency has been consistently improved, with almost no loss of accuracy. Code is available at \url{https://github.com/xugezheng/FAMS}.

\section{Related Work}
\textbf{Algorithmic fairness} Fairness has been attached great importance and widely studied in various applications, such as natural language processing \cite{subramanian-etal-2021-fairness, DBLP:journals/corr/abs-2203-07228, gaut-etal-2020-towards}, natural language generation \cite{DBLP:conf/acl/ShengCNP20, gupta2022mitigating, DBLP:journals/corr/abs-2104-07429}, computer vision \cite{DBLP:conf/cvpr/JungLPM21, DBLP:conf/iccv/LiX21}, and deep learning \cite{DBLP:conf/iccv/ZhaoWR21, DBLP:conf/coling/VuNLJ20}. Then various approaches have been proposed in algorithmic fairness. They typically add fair constraints during the training procedure, such as demographic parity or equalized odds  \cite{hardt2016equality,zemel2013learning,verma2018fairness,jiang2020wasserstein,feldman2015computational,calmon2017optimized}. Apart from this, other fair notions are adopted such as accuracy parity \cite{Sagawa*2020Distributionally,zhao2019conditional}, which requires each subgroup to attain the same accuracy; small prediction variance \cite{li2019fair,li2021ditto}, which ensures small prediction variations among the subgroup; or small prediction loss for all the subgroups \cite{hashimoto2018fairness,martinez2019fairness,balashankar2019fair,zafar2019fairness}. Furthermore, based on the concept of Independence (e.g. demographic parity 
$A\perp\!\!\!\!\perp f(X)$) or conditional independence (e.g. equalized odds $A\perp\!\!\!\!\perp f(X)|Y$ or group sufficiency $A\perp\!\!\!\!\perp Y|f(X)$), another popular line in fair learning is then naturally integrated with information theoretical framework through adding mutual information constraints such as \cite{song2019learning,madras2018learning}. 

\textbf{Understanding fairness-accuracy trade-off} As for the theoretical aspect, \cite{huang2019stable} further investigated the relation of fairness (demographic parity) and algorithmic stability. \cite{dutta2020there} formally justified the inherent trade-off between fairness (w.r.t. demographic parity and equalized odds) and accuracy, whereas the analysis is conducted for the binary sensitive attribute with the population loss. \cite{wang2021understanding} studied the fair-accuracy trade-off in the multi-task learning.

\textbf{Group sufficiency} The fair notion of group sufficiency has recently been highlighted in various real-world scenarios such as health \cite{doi:10.1126/science.aax2342} and crime prediction \cite{chouldechova2017fair,pleiss2017fairness}. Specifically, \cite{pmlr-v97-liu19f} demonstrated that under proper assumptions, group sufficiency can be controlled in the unconstraint learning. However, this conclusion may not necessarily always hold in the overparameterized models with limited samples per subgroup, where \cite{doi:10.1126/science.aax2342,Shui2022FairRL,koh2021wilds} essentially revealed the prediction disparities between the different subgroups in the unconstraint learning. \cite{lee2021fair} recently studied the fair selective classification w.r.t. group sufficiency through an information theoretical framework, while the theoretical guarantee is unknown. In contrast, our proposed lower-level loss within the paper can provably control the generalization error, and the upper-level loss controls the group sufficiency gap. Besides, a close notion to the group sufficiency is the \emph{probability calibration} \cite{guo2017calibration}, which is defined as $\E[Y|f(X)]= f(X)$ in binary classification. 
We will empirically show the probability calibration could be consistently improved within our framework, whereas the analysis on finite samples and its theoretical relation with group sufficiency remains still opening \citep{creager2021environment}.  

\textbf{Bi-level optimization in fairness}  Bi-level optimization seeks to solve problems with a hierarchical structure. Namely, two levels of optimization problems where one task is nested inside another \cite{9638340}. Several ideas related to bi-level optimization have been proposed in the context of fair-learning. 
For instance, we could design a min-max optimization to learn fair representation when considering demographic parity (DP) or equalized odds (EO) \cite{zemel2013learning,song2019learning,zhao2019conditional}.
In this context, a representation function aims to minimize the loss caused by the discriminator in the lower-level. Simultaneously, in the upper-level, a discriminator could be introduced to maximize the loss. Then fair representation could be enforced through the bi-level optimization. Besides, if the accuracy and its variants are tracked as the metrics for each subgroup \cite{gupta2022mitigating}, the bi-level objective could also be deployed in controlling the loss \cite{raghavan2020mitigating} or the prediction variance \cite{li2021ditto}, where the lower-level's goal is to minimize the loss for each subgroup and the upper-level's goal is to estimate the prediction disparities. In our paper, we \emph{theoretically} justified a novel bi-level optimization perspective: controlling group sufficiency and accuracy. Simultaneously, other bi-level optimization and its relevant \emph{meta-learning} algorithms could be further considered in the fair learning such as recurrent based gradient updating \cite{JMLR:v18:17-468}, layer-wise transformation \cite{park2019meta} or implicit gradient based approach \cite{shui2022fair}.

\section{Preliminaries}\label{sec:def}
We assume the joint random variable $(X,Y,A)$ follows an underlying distribution $\D(X,Y,A)$, where $X\in\calX$ is the input, $Y\in\calY$ is the label, and the \emph{scalar} discrete random variable $A\in\calA$ denotes the sensitive attribute (or subgroup index). For instance, $A$ represents gender, race, or age. We also denote $\E[Y|X]$ as the conditional expectation of $Y$, which is essentially a function of $X$. $\E_{A,X}[\cdot]$ is denoted as the expectation on the marginal distribution of $\D(A,X)$. Throughout the paper, we consider binary classification with $\calY=\{0,1\}$. We further define the predictor as a scoring function  $f:\calX\to[0,1]$ that maps the input into a real value in $[0,1]$. It is worth mentioning that in general $f(X) \notin \calY$  since $f(X)$ is continuous. We then introduce group sufficiency and group sufficiency gap.
\begin{definition}[Group sufficiency \cite{barocas-hardt-narayanan,chouldechova2017fair,pmlr-v97-liu19f}] A predictor $f$ satisfies group sufficiency with respect to the sensitive attribute $A$ if~ 
$\E[Y|f(X)] = \E[Y|f(X),A]$.
\end{definition}
Intuitively, given a output score of the predictor $f(X)=\tau$, the conditional expectation of $Y$ is invariant across different subgroups. Namely, conditioning on the specific subgroup $A=a$ does not provide any additional information about the conditional expectation of $Y$. Then we could naturally define group sufficiency gap.

\begin{definition}[Group sufficiency gap \cite{pmlr-v97-liu19f}] The group sufficiency gap of a predictor $f$ is defined as: 
$\textbf{Suf}_f = \E_{A,X}  [\left|\E[Y|f(X)]-\E[Y|f(X),A]\right|]$
\end{definition}

Specifically, $\textbf{Suf}_f$ measures the extent of group sufficiency violation, induced by the predictor $f$, which is taken by the expectation over $(X,A)$. Clearly, $\textbf{Suf}_f = 0$ suggests that $f$ satisfies groups sufficiency and vice versa. For completeness, we also discuss other popular group fairness criteria: demographic parity and equalized odds. 
\begin{definition}[Demographic Parity (DP)] A predictor $f$ satisfies the demographic parity with respect to the sensitive attribute $A$ if: 
$\E[f(X)] = \E[f(X)|A]$
\end{definition}
Demographic Parity (DP), also known as statistical parity or independence rule, emphasizes that the expectation of the output score $f(X)$ is independent of $A$. \cite{barocas-hardt-narayanan,chouldechova2017fair} further revealed that if $A \not\!\perp\!\!\!\perp Y$, group sufficiency and demographic parity could not be simultaneously achieved. 

\begin{definition}[Equalized Odds (EO) \citep{hardt2016equality}] A predictor $f$ satisfies  the equalized odds with respect to $A$ if:
$\E[f(X)|Y] = \E[f(X)|Y,A]$
\end{definition}
Equalized odds (EO) emphasizes the conditional expectation of output $f$ is invariant w.r.t. $A$, given the ground truth $Y$. \cite{barocas-hardt-narayanan,pleiss2017fairness} reveal that if $\D(X,Y,A)>0$ and $A \not\!\perp\!\!\!\perp Y$, group sufficiency and equalized odds can not both hold. 

The analysis reveals a general \emph{incompatibility} between group sufficiency and DP/EO when $A \not\!\perp\!\!\!\perp Y$, which often occurs in practice. Besides, DP/EO based criteria generally suffers the well-known fair accuracy trade-off \cite{song2019learning}: enforcing the fair constraint degrades the prediction performance. This paper depicts that under the criteria of group sufficiency, these objectives could be both encouraged.
 
\section{Upper bound of group sufficiency gap}
To derive the theoretical results, we first introduce the group Bayes predictor.
\begin{definition}[$A$-group Bayes predictor]\label{def:g_bayes}
The $A$-group Bayes predictor $f_{A}^{\text{Bayes}}$ is defined as:
$f_{A}^{\text{Bayes}}(X) = \E[Y|X,A]$
\end{definition}
The $A$-group Bayes predictor is associated with the underlying data  distribution $\D(X, Y, A)$. Given the fixed realization $X=x, A=a$, we have $f_{A=a}^{\text{Bayes}}(x) = \E[Y|X=x, A=a]$, which suggests the ground truth conditional data generation of subgroup $A=a$. By adopting $f_{A=a}^{\text{Bayes}}(x)$, we could derive the upper bound of group sufficiency gap w.r.t. any predictor $f$:

\begin{theorem} \label{thm:uff_upp} Group sufficiency gap $\textbf{Suf}_f$ is upper bounded by:~~$\textbf{Suf}_f \leq 4\E_{A,X}[|f-f_{A}^{\text{Bayes}}|]$\\
Specifically, if $A$ takes finite value ($|\calA|<+\infty$) and follows uniform distribution with $\D(A=a)=1/|\calA|$. Then the group sufficiency gap is further simplified as:
\[
\textbf{Suf}_f \leq \frac{4}{|\calA|}\sum_{a}\E_{X}[|f-f_{A=a}^{\text{Bayes}}||A=a]
\]
\end{theorem}
The proof is inspired by \cite{pmlr-v97-liu19f}. Specifically, Theorem \ref{thm:uff_upp} reveals that the upper bound of group sufficiency gap depends on the discrepancy between the predictor $f$ and $A$-group Bayes predictor $f_{A}^{\text{Bayes}}(X)$. Namely, given different subgroups $A=a$, the optimal predictor $f$ ought to be closed to all the group Bayes predictors $f_{A=a}^{\text{Bayes}}(X)$, $\forall a\in\calA$.

\textbf{Underlying assumption} Theorem \ref{thm:uff_upp} also reveals underlying assumptions w.r.t. the data generation distribution $\D(X,Y,A)$ for achieving a small group sufficiency gap. If $f_{A}^{\text{Bayes}}$ for each subgroup $A=a$ are quite similar, then minimizing the upper bound yields a small group sufficiency gap $\textbf{Suf}_f$. For example, consider the extreme scenario \textbf{if} the $A$-group Bayes predictors are identical w.r.t. $A$, 
$\E[Y|X,A=a] = \E[Y|X], \forall a\in\calA $, where $\E[Y|X]$ is the conventional Bayes predictor defined on the marginalized distribution $\D(X,Y)$. The upper bound recovers the difference between the predictor $f$ and standard Bayes predictor. If we use a probabilistic framework to approximate predictor $f(X) \approx \E[Y|X]$ (i.e, training the entire dataset without any fair constraint), both group sufficiency gap and prediction error (since Bayes predictor is optimal) will be small, which is consistent with \citep{pmlr-v97-liu19f}. 
On the contrary, if $A$-group Bayes predictors are completely arbitrary with high variance for $A$, both group sufficiency gap and prediction error are large and it would be impossible for an informative prediction. 

\section{Principled Approach}
Based on the upper bound, we propose a principled approach to learn the predictor that achieves both small generalization error and group sufficiency gap.

\subsection{Upper bound in randomized algorithm}
To establish the theoretical result, we consider a randomized algorithm that learns a predictive-distribution $Q$ over scoring predictors from the data. For instance, if we consider Bayes framework, the predictor is drawn from the posterior distribution $\tilde{f}\sim Q$. In the inference, the predictor's output is formulated as the expectation of the learned predictive-distribution $Q$:
$f(X) = \E_{\tilde{f}\sim Q} \tilde{f}(X)$.

\begin{wrapfigure}{r}{5cm}
    \centering
    \includegraphics[width=47mm]{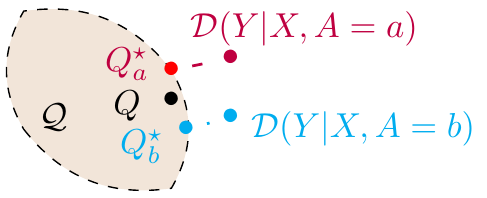}
    \caption{Illustration of optimization and approximation term. In the binary subgroup $\calA=\{a,b\}$, 
    the optimization term is to find $Q\in\mathcal{Q}$ to minimize the discrepancy between $(Q_a^{\star},Q_b^{\star})$. 
    The approximation term is solely based on the distribution family $\mathcal{Q}$ (\textcolor{brown}{brown} region). If the predefined $\mathcal{Q}$ has a rich expressive power, the 
    approximation is treated as a small constant. }
    \label{fig:error_decomp}
\end{wrapfigure}

In practice, it is infeasible to optimize over all the possible distributions. Then we should restrict the predictive-distribution $Q$ within a distribution family $Q \in \mathcal{Q}$ such as Gaussian distribution. We also denote $Q_a^{\star}\in\mathcal{Q}$ as the optimal prediction-distribution w.r.t. $A=a$ under binary cross-entropy loss within the distribution family $\mathcal{Q}$:~ $Q_a^{\star} := \text{argmin}_{Q_a\in\mathcal{Q}}~\E_{\tilde{f}_a \sim Q_a} ~\calL_{a}^{\text{BCE}}(\tilde{f}_a)$. In generally $Q^{\star}_{a} \neq f^{\text{Bayes}}(x,A=a)$, since the distribution family $\mathcal{Q}$ is only the subset of all possible distributions (shown in Fig.~\ref{fig:error_decomp}). We then extend the upper bound in the randomized algorithm.
\begin{corollary}\label{eq:random_bound}
The group sufficiency gap $\textbf{Suf}_f$ in randomized algorithm w.r.t. learned predictive-distribution $Q$ is upper bounded by:
\[\textbf{Suf}_{f} 
     \leq \frac{2\sqrt{2}}{|\calA|}[ \sum_{a}\underbrace{\sqrt{\text{KL}(Q_a^{\star}\|Q)}}_{\text{Optimization}} 
      +\underbrace{\sqrt{\text{KL}(Q_a^{\star}\|\D(Y|X,A=a))}}_{\text{Approximation}}] \]
\end{corollary}
Where $\text{KL}$ is the Kullback–Leibler divergence. Corollary \ref{eq:random_bound} further reveals that the upper bound is  decomposed into two terms, showing in Fig.\ref{fig:error_decomp}.

\textbf{Optimization term}~~The optimization term is the average KL divergence between the learned distribution $Q$ and optimal predictive-distribution $Q^{\star}_{a}$ for each subgroup $A=a$.
Minimizing the optimization term implies that the learned distribution $Q$ will be both fair and informative for the prediction, because it aims to minimize the upper bound of the group sufficiency gap $\textbf{Suf}_{f}$ and be close to the optimal predictive-distribution w.r.t. each $A=a$. 

\textbf{Approximation term}~~The approximation term is the average KL divergence between the optimal distribution $Q^{\star}_a$ and the underlying data generation distribution. 
Given the distribution family $\mathcal{Q}$, it is a unknown constant. Besides, if the distribution family $\mathcal{Q}$ has a rich expressive power such as deep neural-network, the approximation term will be small \citep{shalev2014understanding}. However, an extreme large distribution family $\mathcal{Q}$ could simultaneously yield a potential overfitting on finite samples. In this paper, the neural network is adopted and the approximation term is assumed to be a small constant. Thus, controlling $\textbf{Suf}_{f}$ implies minimizing the optimization term.

\subsection{Challenge in learning limited samples}
In practice, we only have access to finite or even limited samples in each subgroup, rather than the underlying distribution $\D$. We denote $S_a=\{(x^{a}_i,y^{a}_i)\}_{i=1}^{m}$ as the observed data w.r.t. subgroups $A=a$, which are i.i.d. samplings from the underlying  distribution $\D(x,y|A=a)$. We also denote the \textbf{empirical} binary cross entropy loss w.r.t $A=a$ as: $\hat{\calL}_{a}^{\text{BCE}}(\tilde{f}) = \frac{1}{m}\sum_{i=1}^m -[y^{a}_i\log(\tilde{f}(x^{a}_i)) + (1-y^{a}_i)\log(1-\tilde{f}(x^{a}_i))]$.
Then a straight approach is to minimize the empirical term $\hat{Q}_{a}^{\star}$:
\begin{align}\label{eq:erm}
    \hat{Q}_a^{\star} = \text{argmin}_{Q_a\in\mathcal{Q}}~\E_{\tilde{f}_a \sim Q_a} ~\hat{\calL}_{a}^{\text{BCE}}(\tilde{f}_a)
\end{align}
Then $Q$ is updated through minimizing the average KL-divergence: $\sum_{a}\text{KL}(\hat{Q}^{\star}_{a}\|Q)$ from learned $\hat{Q}_{a}^{\star}$. However, this idea generally \emph{does not work} in our setting, because each subgroup contains \textbf{limited} number of samples. Therefore, a straight minimization leads to overfitting for each subgroup and generalization error $\E_{\tilde{f}\sim \hat{Q}^{\star}_a} ~\calL_{a}^{\text{BCE}}(\tilde{f})$ is quite large, showing in Fig.~\ref{fig:amazon_lambda}.
\subsection{$Q$ as an informative prior}\label{sec:bilevel}
We have demonstrated that $Q$ can achieve both fair and informative prediction. Therefore, we regard $Q$ as a \emph{prior} information for minimizing the loss, yielding a bilevel objective.
\begin{align*}
     & \min_{Q\in\mathcal{Q}} \frac{1}{|\calA|}\sum_{a}\text{KL}(\overline{Q}_a^{\star}\|Q) \tag*{(\textcolor{cyan}{Upper-level})}\\
     & \text{s.t.}~~\overline{Q}_a^{\star} = \text{argmin}_{Q_a\in\mathcal{Q}} \{\E_{\tilde{f}_a\sim Q_a} ~\hat{\calL}_{a}^{\text{BCE}}(\tilde{f}_a) + \lambda \text{KL}(Q_a\|Q)\}, \forall a\in\calA \tag*{(\textcolor{brown}{Lower-level})}
\end{align*}
Where $\lambda>0$ is the hyper-parameter. The proposed loss is a typical bilevel optimization. \textcolor{brown}{(1)} In the lower-level, we aim to learn $\overline{Q}_a^{\star}$ for each $a\in\mathcal{A}$. Different from Eq.~(\ref{eq:erm}), the loss in lower-level adds a regularization term $\text{KL}(Q_a\|Q)$ as an informative prior in learning $\overline{Q}_a^{\star}$, given a fixed predictive-distribution $Q$. Moreover, Theorem \ref{thm:pac_bayes} formally justified that optimizing the lower-level loss is to minimize the upper bound of the generalization error. \textcolor{cyan}{(2)} In the upper-level, $Q$ is updated through minimizing the average KL divergence between different $\overline{Q}_a^{\star}$, which controls the upper bound of $\textbf{Suf}_f$.

\begin{theorem}[Generalization error bound] \label{thm:pac_bayes}Supposing that datasets $\{S_a\}_{a=1}^{|\calA|}$ with $S_a=\{(x^{a}_i,y^{a}_i)\}_{i=1}^{m}$ are i.i.d. sampled from $\D(x,y|A=a)$, the binary cross entropy (BCE) loss is upper bounded by $L$, $Q_a\in\mathcal{Q}$ is any learned distribution from dataset $S_a$ and $Q\in\mathcal{Q}$ is any distribution. Then with high probability $\geq 1-\delta$ with $\forall\delta\in (0,1)$, we have:
\begin{align*}
\begin{small}
    \frac{1}{|\calA|}\sum_{a} \E_{\tilde{f}_a \sim Q_a}  \calL_{a}^{\text{BCE}}(\tilde{f}_a) \leq  \underbrace{\frac{1}{|\calA|}\sum_{a} \E_{\tilde{f}_a \sim Q_a}  \hat{\calL}_{a}^{\text{BCE}}(\tilde{f}_a)}_{(1)} 
      + \underbrace{\frac{L}{\sqrt{|\calA|m}}\sum_{a}\sqrt{\text{KL}(Q_a\|Q)}}_{(2)}+\underbrace{L\sqrt{\frac{\log(1/\delta)}{|\calA|m}}}_{(3)}
\end{small}
    \end{align*}
\end{theorem}
\textbf{Discussions}~ The proof is inspired by PAC-Bayes theorem such as \cite{pentina2014pac,amit2018meta,chen2021generalization}. Secpficially, Theorem \ref{thm:pac_bayes} reveals the generalization error in the lower-level is upper bounded by three terms. \textcolor{blue}{(a)} Term (1) is the average empirical prediction error, which corresponds to the first term in the lower-level loss. \textcolor{blue}{(b)} Term (2) indicates the average KL-divergence between the learned subgroup distribution $Q_a$ and the prior distribution $Q$, which corresponds to the second term in the lower-level loss. The combination of term (1-2) recovers the averaged lower-level loss w.r.t. $A$. \footnote{In Theorem \ref{thm:pac_bayes}, the differences are in the square norm of KL divergence and setting the specific hyper-parameter: $\lambda=L\sqrt{|\calA|/m}$.} 
Thus optimizing the lower-level loss could control the generalization error. \textcolor{blue}{(c)} When the confidence $\delta$ is fixed, term (3) will converge if $|\calA|m\to +\infty$. Moreover, even if $m$ (the sample size in each subgroup) is quite small, a sufficient large number of subgroups $|\calA|$ can also ensure the convergence of term (3).

For the sake of simplicity, we assumed the identical samples size $m$ in each subgroup $S_a$, while the theoretical result can be extended to subgroups with different samples $m_a$. 

\subsection{Practical Implementations}
In this section, we develop a practical learning algorithm that can be applied to a wide range of differentiable and parametric models, including neural networks.

\paragraph{Parametric models} We choose the Isotropic Gaussian distribution (with diagonal covariance matrix) as the distribution family $\mathcal{Q}$, where the mean and covariance are set as $d$-dimensional parameter. Thus we need to learn the parameter $(\boldsymbol\theta,\boldsymbol\sigma)$ for fair and informative $Q\in\mathcal{Q}$. As for the subgroup $A=a$, we learn parameters $(\boldsymbol\theta_{a},\boldsymbol\sigma_{a})$ for $\overline{Q}^{\star}_a\in\mathcal{Q}$. It is worth mentioning that the Isotropic Gaussian distribution is selected for its computational efficiency in the optimization. We can use any distribution as long as the density function is differentiable with respect to the parameters. 

For the single predictor $\tilde{f}$, we use parametric neural-network models and assume $\tilde{f}$ is parameterized by a $d$-dimensional vector ${\bf{w}}\in \R^{d}$, denoted as $\tilde{f}_{\bf{w}}$.  Then $\tilde{f}_{\bf{w}} \sim Q$ is equivalent to sampling the model parameter $\bf{w}$ from the predictive-distribution $Q$: $\mathbf{w} \sim \calN(\boldsymbol\theta,\boldsymbol\sigma^2) = \prod_{i=1}^{d}\calN(\boldsymbol\theta[i],\boldsymbol\sigma^2[i])$.
Since $Q$ is Isotropic Gaussian, each element $i$ in the parameter ${\bf w}[i]$ follows a 1-dimensional Gaussian. Following the same line, $\tilde{f}_{{\bf{w}}_a} \sim \overline{Q}^{\star}_{a}$ can be modeled analogously:
$\mathbf{w}_a \sim \calN(\boldsymbol\theta_{a},\boldsymbol\sigma_{a}^2) = \prod_{i=1}^{d} \calN(\boldsymbol\theta_{a}[i],\boldsymbol\sigma^2_{a}[i])$. As a result, learning the distribution $Q$ is equivalent to learning parameter $(\boldsymbol\theta,\boldsymbol\sigma)$ in the bilevel objective.

\paragraph{Gradient Estimation} Based on the previous setting, we aim to optimize the bilevel objective to obtain the parameter of $Q$: $(\boldsymbol\theta,\boldsymbol\sigma)$. We use stochastic gradient descent (SGD) to optimize the parameters. In the lower-level, the loss in Sec.~\ref{sec:bilevel} is composed by the empirical prediction error and KL divergence term. The KL divergence has a closed form that can be differentiated efficiently.
Specifically, since $Q$ and the subgroup specific $\overline{Q}^{\star}_{a}$ are factorized Gaussian, the KL divergence takes a simple closed form and the gradient can be easily calculated:
$\text{KL}(\overline{Q}^{\star}_a\|Q) = \frac{1}{2}\sum_{i=1}^d\left\{\log\frac{\boldsymbol\sigma^2_{a}[i]}{\boldsymbol\sigma^2[i]}+ \frac{\boldsymbol\sigma^2_{a}[i]+(\boldsymbol\theta_{a}[i]-\boldsymbol\theta[i])^2}{\boldsymbol\sigma^2[i]} - 1\right\}$.

\paragraph{Re-parametrization trick} 
As for the prediction error $\E_{\tilde{f}_{{\bf{w}}_{a}}\sim Q_a} ~\hat{\calL}_{a}^{\text{BCE}}(\tilde{f}_{{\bf{w}}_{a}})$, the term $\hat{\calL}_{a}^{\text{BCE}}(\tilde{f}_{{\bf{w}}_{a}})$  is non-linear for $\bf{w}$, rendering the expectation intractable in the computation. To this end, we adopt the re-parameterization trick \citep{kingma2013auto,blundell2015weight} in computing the gradient w.r.t. the expectation term. The trick is based on describing the Gaussian distribution ${\bf{w}}_a\sim\calN({\boldsymbol\theta}_{a},{\boldsymbol\sigma}^2_{a})$ as first
drawing $\boldsymbol\epsilon\sim\calN(\mathbf{0},\mathbf{I})$ and then applying the deterministic function ${\bf{w}}_a({\boldsymbol\theta}_{a},{\boldsymbol\sigma}_{a}) = \boldsymbol\theta_{a} + \boldsymbol\sigma_{a} \odot \boldsymbol\epsilon$ ($\odot$ is element-wise product) to approximate the sampling. Then the gradient term can be estimated as: $\nabla_{({\boldsymbol\theta}_{a},{\boldsymbol\sigma}_{a})} \E_{{\bf{w}}_{a}\sim N({\boldsymbol\theta}_{a},{\boldsymbol\sigma}_{a})} ~\hat{\calL}_{a}^{\text{BCE}}(\tilde{f}_{{\bf{w}}_{a}}) = \nabla_{(\boldsymbol\theta_{a},\boldsymbol\sigma_{a})} \E_{\boldsymbol\epsilon\sim\calN(\mathbf{0},\mathbf{I})} \hat{\calL}_{a}^{\text{BCE}}(\tilde{f}_{{\bf{w}}_{a}({\boldsymbol\theta}_{a},{\boldsymbol\sigma}_{a})})$, where the expectation can be approximated by Monte-Carlo sampling w.r.t. $\boldsymbol{\epsilon}$. For a fixed sample $\boldsymbol\epsilon$, the gradient can be computed through backpropagation. In the upper-level, the KL divergence has a closed form, thus it is easy to update the parameter of $Q$ through backpropagation.

\paragraph{Proposed Algorithm} Based on the analysis, the algorithm is shown in Algorithm.~\ref{bi_level_main} for solving the bilevel objective in Sec.~\ref{sec:bilevel}. Specifically, we adopt the alternating optimization. Namely, in the lower-level, we fix $Q$ and optimize the subgroup specific predictor $\overline{Q}^{\star}_{a}$ through SGD. Then in the upper-level, we fix the learned $\overline{Q}^{\star}_{a}$ and update $Q$. Since we may face many subgroups, at each training epoch, we randomly sample a subset $\calA^{'}$ such that $|\calA^{'}|\ll |\calA|$ for the memory saving.

\begin{algorithm}[!t]
		\caption{Fair and Informative Learning for Multiple Subgroups (FAMS)}
		\begin{algorithmic}[1] 
        \STATE {\bfseries Input:} Parameters w.r.t. distribution $Q$:$(\boldsymbol\theta,\boldsymbol\sigma^2)$, datasets $\{S_a\}$, $a\in\calA$.
        \FOR{Sampling a subset of $\{S_a\}$, where $a\in\calA^{\prime}\subseteq \calA$}
        \STATE \textcolor{brown}{\#\#\#~Solving the lower-level~\#\#\#}
        \STATE Fix $Q$, optimizing the loss w.r.t. $Q_a=\calN(\boldsymbol\theta_{a},\boldsymbol\sigma^2_{a})$ through SGD for each $a\in\calA^{\prime}$
        
        $\quad\quad\quad\quad\quad\quad\quad\quad\quad \E_{\tilde{f}_{{\bf w}_a}\sim Q_a} ~\hat{\calL}_{a}^{\text{BCE}}(\tilde{f}_{{\bf w}_a}) + \lambda \text{KL}(Q_a\|Q)$
        
        \STATE Obtaining the solution $\overline{Q}^{\star}_{a}$, $a\in\calA^{\prime}$.
        \STATE \textcolor{cyan}{\#\#\#~Solving the upper-level~\#\#\#}
        \STATE Fix $\overline{Q}^{\star}_{a}$ with $a\in\calA^{\prime}$, optimizing the loss w.r.t. $Q$ through SGD:
        $\frac{1}{|\calA^{\prime}|}\sum_{a}\text{KL}(\overline{Q}_a^{\star}\|Q) $    
        \STATE Obtaining updated parameter $(\boldsymbol\theta,\boldsymbol\sigma^2)$ in $Q$
        \ENDFOR
    \STATE \textbf{Return:}~~Parameter of distribution $Q$: $(\boldsymbol\theta,\boldsymbol\sigma^2)$
        \end{algorithmic}
        \label{bi_level_main}
\end{algorithm}

\paragraph{Inference} In the inference, we use the Monte-Carlo method to sample the weights of the neural network from distribution ${\bf{w}} \sim \calN(\boldsymbol\theta,\boldsymbol\sigma^2)$, then averaging the output w.r.t. different sampled weights to approximate
$f(x) =\E_{\tilde{f}_{\bf{w}}\sim Q}\tilde{f}_{\bf{w}}(x)$

\section{Experiments}
\subsection{Experimental Setup}
\paragraph{Dataset: Amazon review} We adopt Amazon review dataset \citep{ni2019justifying,koh2021wilds}, which aims to predict the sentiment (classification) from the review. The datasets consist of large-scale users. Each user has limited number of reviews, ranging from 75 to 400. The \emph{user} is then treated as a subgroup, and it has been observed that standard training can lead to prediction disparities in several users.\citep{koenecke2020racial}.The experiment is adapted from the protocol of \citep{koh2021wilds}. Specifically, we convert the original review score (ranging from 1-5) to the binary label: the positive review (score $\geq 4$) and negative review (score $\leq 3$). We draw and then fix 200 users from the original dataset, which includes the training, validation, and test sets. In the implementation, we first adopt DistilBERT \citep{sanh2019distilbert} to learn the embedding with dimension $\R^{768}$. Then we adopt  $\tilde{f}_{\bf{w}}$ and $\tilde{f}_{\bf{w}_a}$ as the four-layer fully connected neural network, where $ {\bf w}\sim Q$ and  $ {\bf w}_a\sim Q_a$. Additional experimental details are delegated to the Appendix. 

\paragraph{Dataset: Toxic Comments} We also use the toxic comment dataset \cite{borkan2019nuanced} to predict the text comment being toxic or not, which has shown the significant performance degradation on specific sub-populations. Following \cite{borkan2019nuanced}, we select \emph{race} as sensitive attribute, which includes Black, White, Asian and Latino \& others (4 subgroups). We also follow the same setting as the original dataset \cite{koh2021wilds}, which has the separate training, validation, and test sets. Since toxic comments are marked by multiple annotators, we decide that the comment is toxic if it is marked by at least half of the annotators. In the implementation, we adopt the DistilBERT \cite{sanh2019distilbert} as the embedding with output dimension $\R^{768}$. Then we also adopt $\tilde{f}_{\bf{w}}$ and $\tilde{f}_{\bf{w}_a}$ as the four-layer fully connected neural network, with the same network structure as Amazon. Additional details are delegated to the Appendix. 

\paragraph{Baselines} We compare with following baselines. (1) ERM: training a deep model without considering the sensitive attribute. (2) SNN. Since we adopt randomized algorithms in the paper, we additionally compare the stochastic neural network through the vanilla training from the whole dataset. Namely, we find a predictive-distribution $Q$ to minimize  $\frac{1}{|\calA|}\sum_{a}\E_{f \sim Q}\hat{\calL}_{a}^{\text{BCE}}(f)$.  (3) EIIL \cite{creager2021environment} proposed an IRM based approach to promote the group sufficiency. (4) FSCS \cite{lee2021fair} adopted the conditional mutual information constraint $I(A,Y|f(X))$ to promote the sufficiency. (5) DRO \cite{Sagawa*2020Distributionally}. A re-weighting approach to assign the importance of the task. Indeed, DRO does not provably guarantee group sufficiency, while it encourages identical losses. All the experiments are repeated five times. 

\paragraph{Computing $\textbf{Suf}_{f}$} Since $f(X)$ is continuous, the group sufficiency gap is calculated by splitting the output of predictor into multiple intervals in $[0,1]$ and computing the conditional expectation within each interval, as detailed in Appendix. 

\begin{figure}[t]
\centering
  \subfigure[Accuracy-$\textbf{Suf}_{f}$]{\label{fig:acc_amazon}\includegraphics[width=40mm]{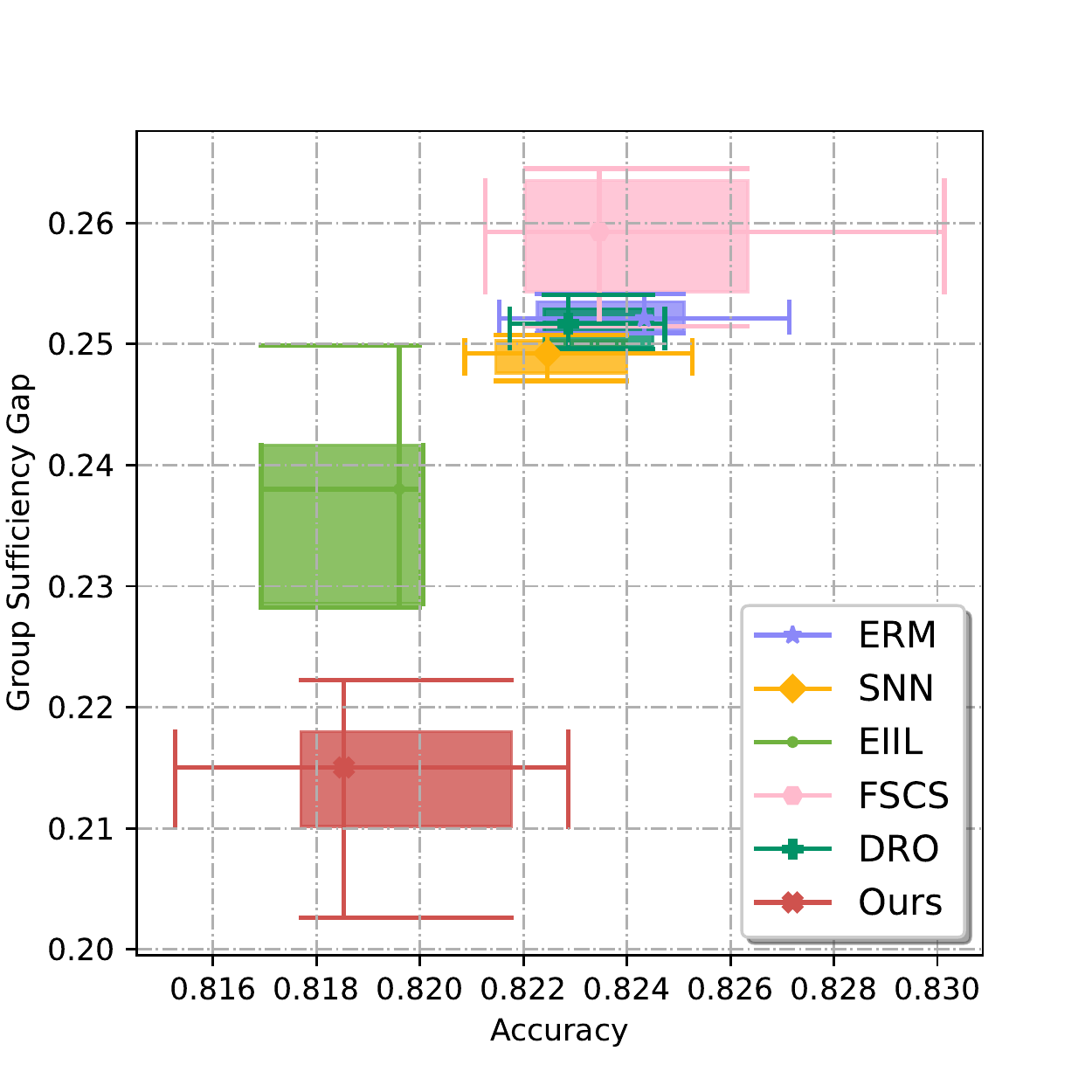}}
  \quad
  \subfigure[$\textbf{Suf}_{f}$ on each subgroup]{\label{fig:suff_amazon}\includegraphics[width=40mm]{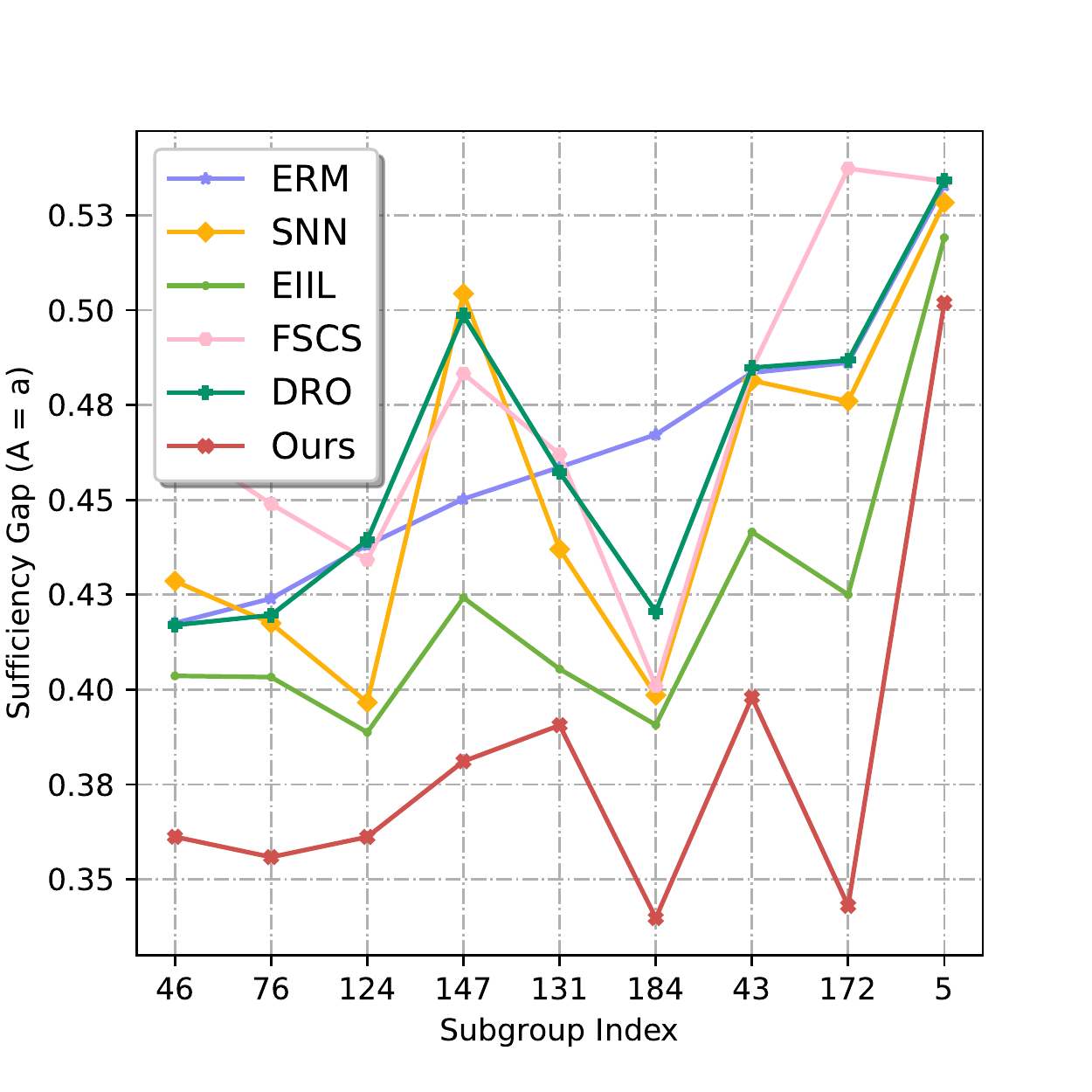}}
  \quad
  \subfigure[Probability Calibration]{\label{fig:cal_amazon}\includegraphics[width=40mm]{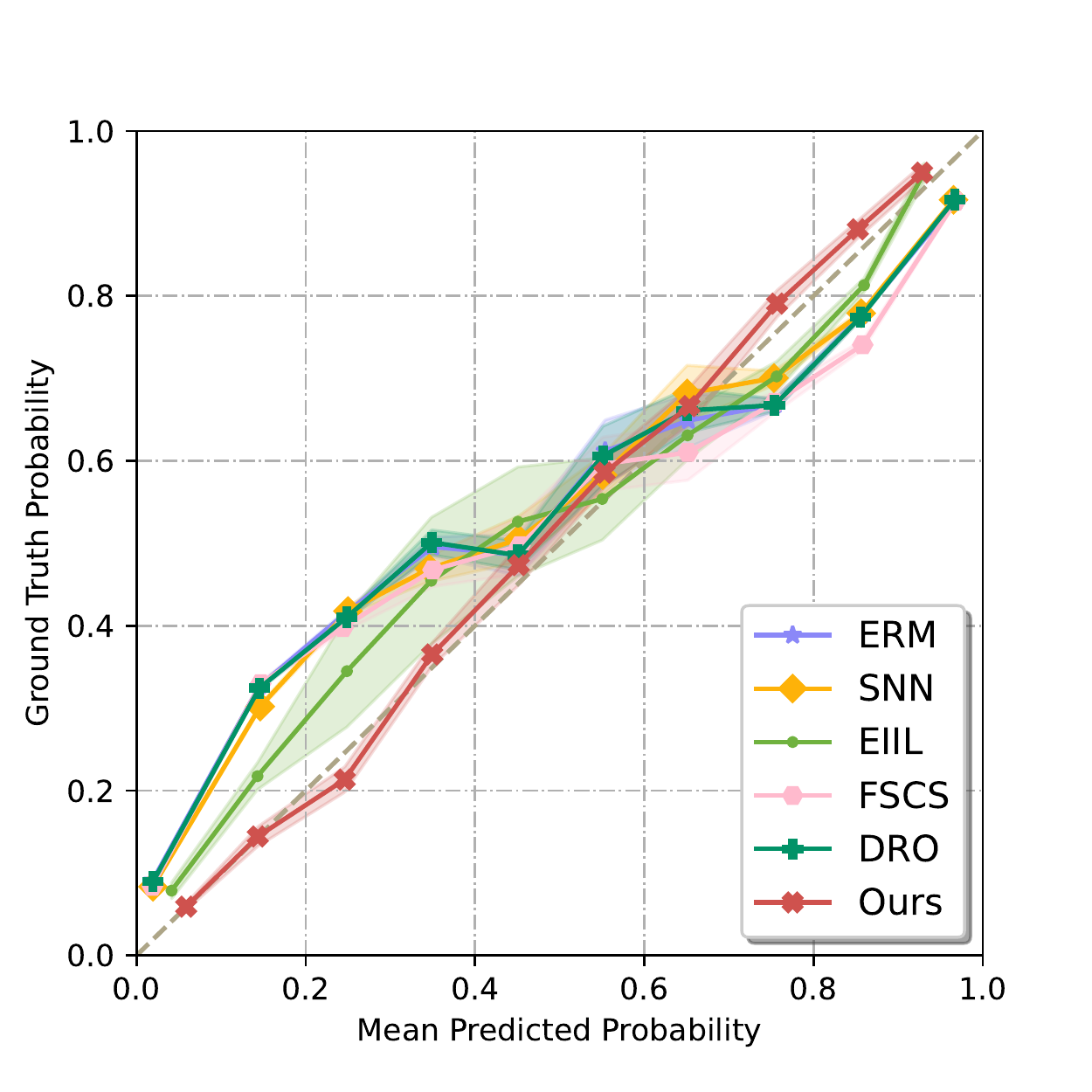}}
  \caption{Amazon Review dataset. (a) Boxplot of accuracy and group sufficiency gap $\textbf{Suf}_f$ with 5 repeats: median, 75th percentile and minimum-maximum value. (b) Group sufficiency gap on subgroup $A=a$, which is the difference between $\E[Y|f(X)]$ and $\E[Y|f(X),A=a]$. We visualize the top-9 users' group sufficiency gap in ERM, whereas the result for all users is delegated to the Appendix.  (c) Probability calibration curve over 5 repeats with mean and standard deviation. i.e $(f(X), \E[Y|f(X)])$. The proposed approach demonstrated a consistently improved probability calibration.}\label{fig:amazon}
\end{figure}

\begin{figure}[t]
\centering
  \subfigure[Accuracy-$\textbf{Suf}_{f}$]{\label{fig:acc_toxic}\includegraphics[width=40mm]{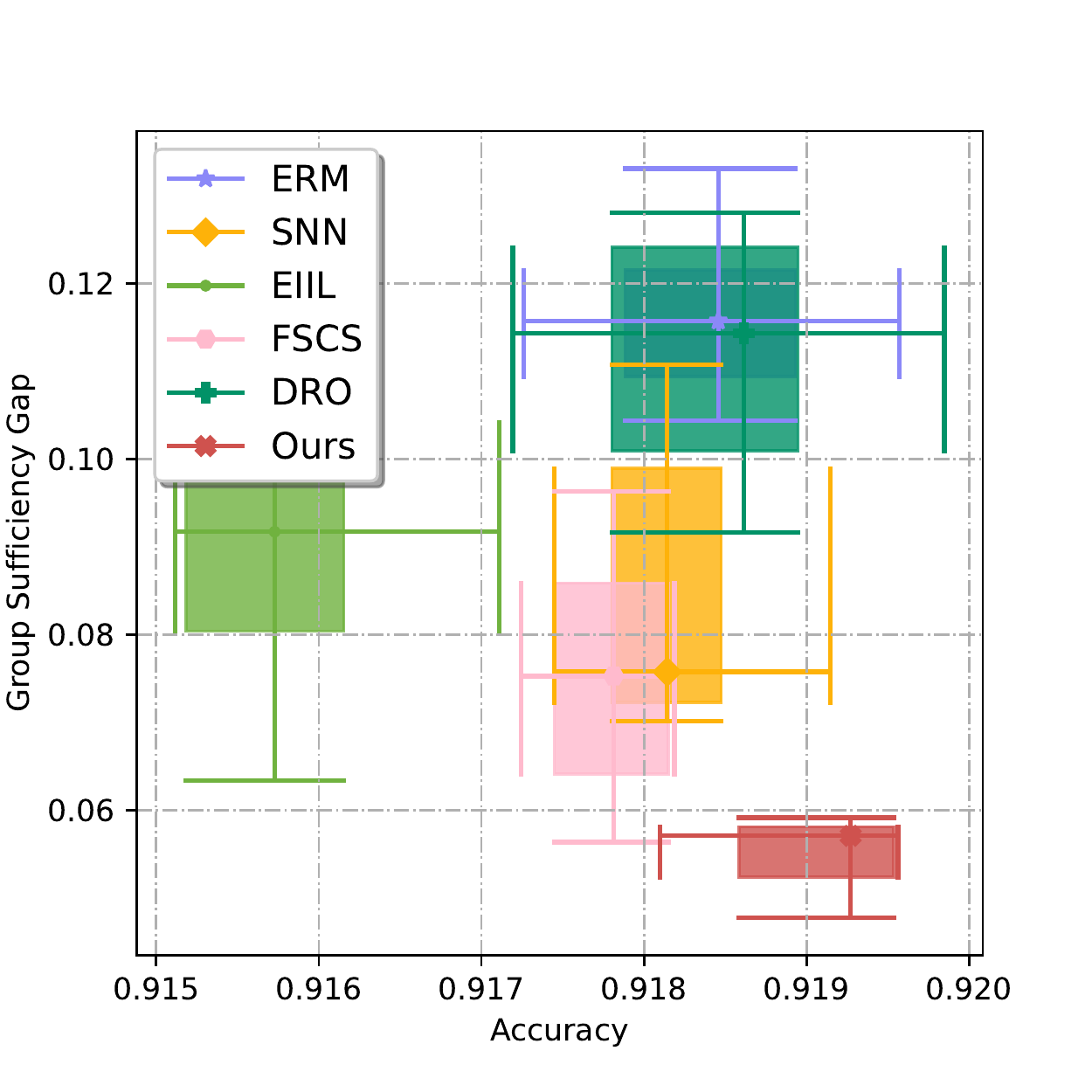}}
  \quad
  \subfigure[$\textbf{Suf}_{f}$ on each subgroup]{\label{fig:suff_toxic}\includegraphics[width=40mm]{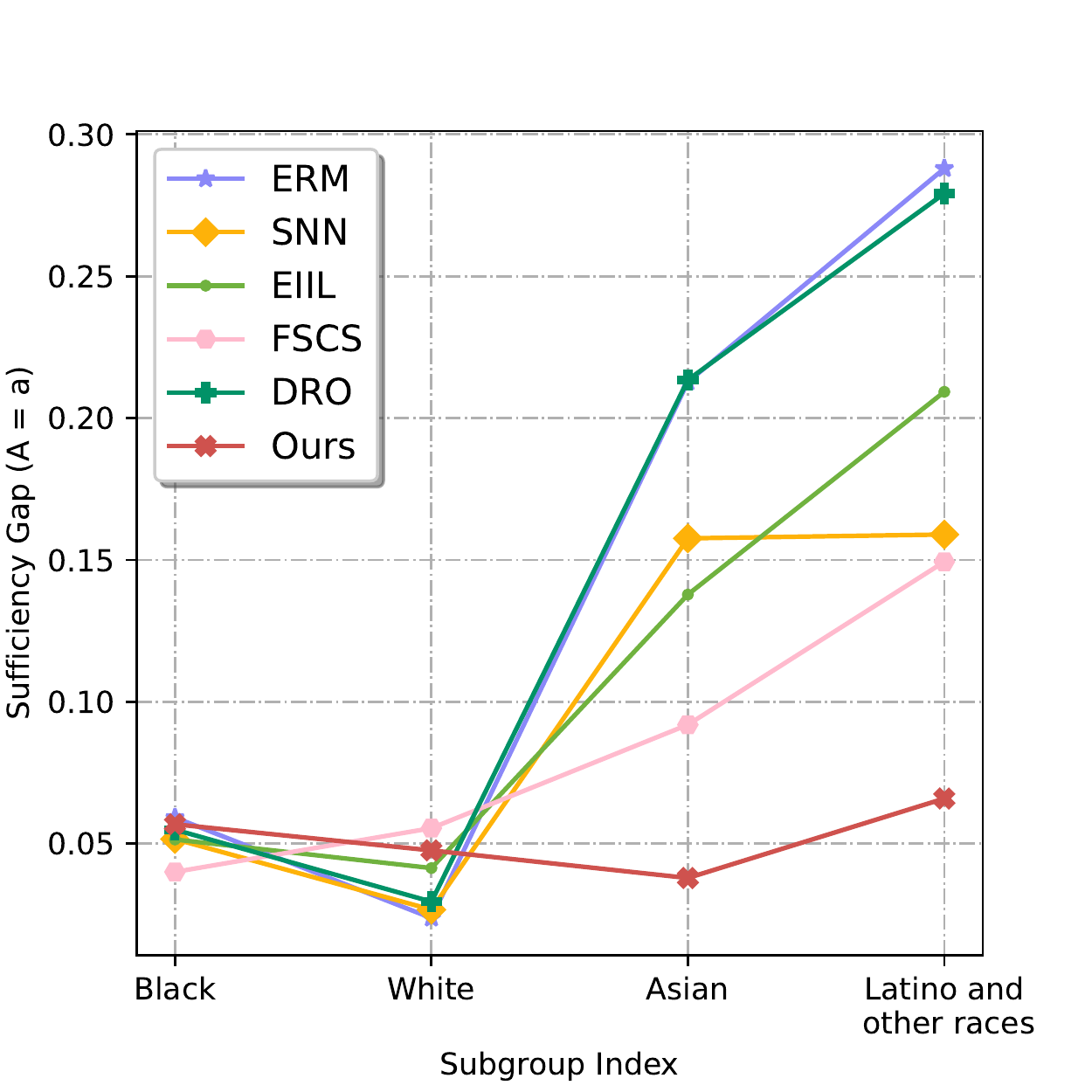}}
  \quad
  \subfigure[Probability Calibration]{\label{fig:cal_toxic}\includegraphics[width=40mm]{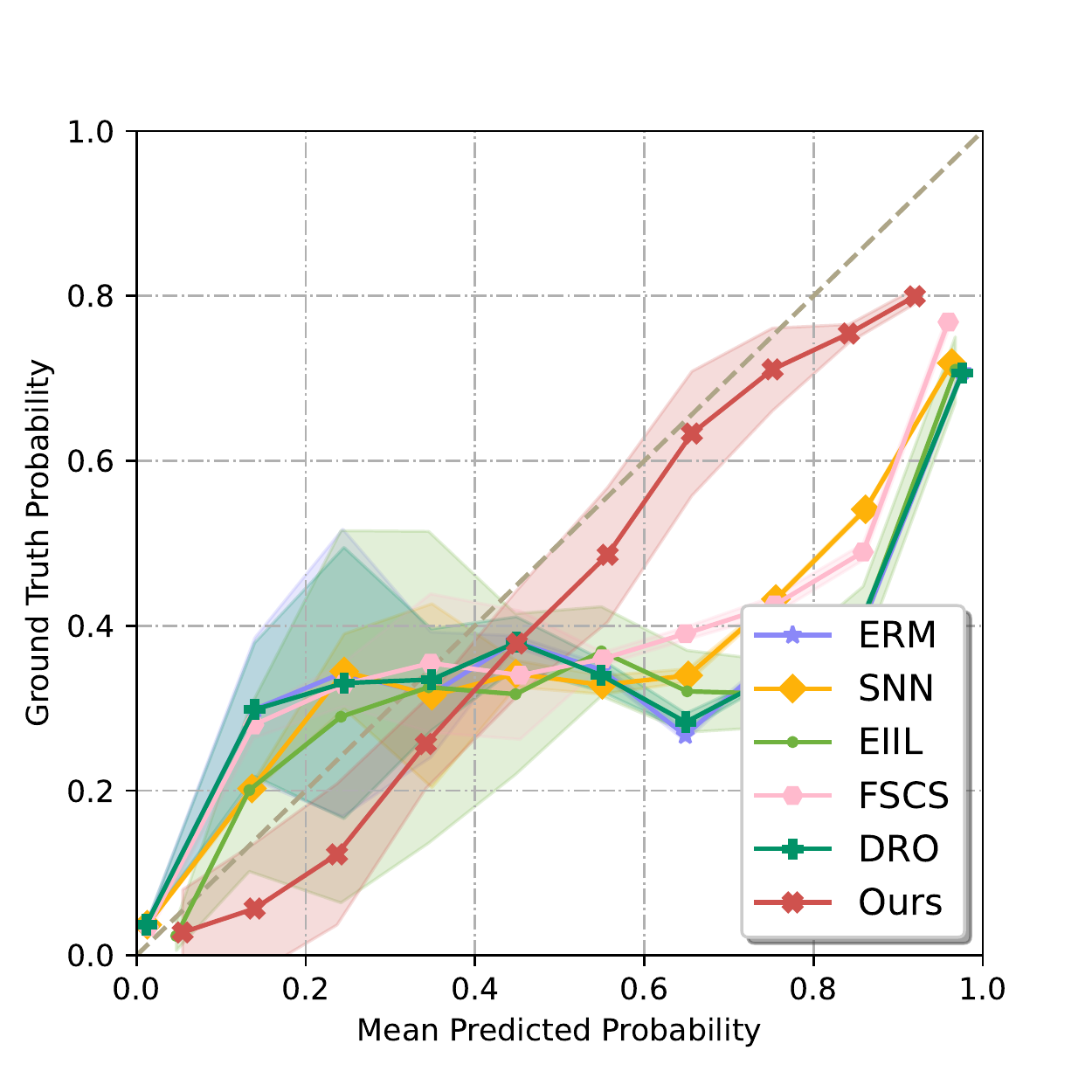}}
  \caption{Toxic dataset. (a) Boxplot of accuracy and group sufficiency gap $\textbf{Suf}_f$ with 5 repeats: median, 75th percentile and minimum-maximum value. (b)  Group sufficiency gap on the specific subgroup $A=a$, which is the difference between $\E[Y|f(X)]$ and $\E[Y|f(X),A=a]$. (c) Probability calibration over 5 repeats with mean and standard deviation. i.e $(f(X), \E[Y|f(X)])$, where the probability calibration for each subgroup is delegated to Appendix. }\label{fig:toxic}
\end{figure}

\subsection{Experimental Results}
We visualize the results in Fig.~\ref{fig:amazon} for Amazon review product and Fig.~\ref{fig:toxic} for toxic comment. 

\textbf{Accuracy and Fairness}~The accuracy and group sufficiency gap are depicted in Fig.~\ref{fig:acc_amazon} and Fig.~\ref{fig:acc_toxic}. In Amazon review, the accuracy in the proposed approach has a slight decrease, compared with ERM. While the group sufficiency gap has improved by $3.0\%$, showing a significant improvement in the fairness. In toxic comments, the accuracy in proposed approach is nearly identical to the baseline, whereas group sufficiency gap has been significantly improved by $3.0$-$3.5\%$.

\textbf{Group sufficiency gap on the specific subgroup}~To gain better understandings of group sufficiency gap, we visualize group sufficiency gap on specific subgroup $A=a$, i.e the discrepancy between the $\E[Y|f(X)]$ (conditional expectation on the entire data) and $\E[Y|f(X),A=a]$ (conditional expectation on subgroup $A=a$). I.e,  $\E_{X}\left[|\E[Y|f(X)]-\E[Y|f(X),A=a]|\right]$. 

In Amazon review dataset, we visualize the top-9 users' sufficiency gap in ERM, as shown in Fig. \ref{fig:suff_amazon}, where the gap of entire users is delegated to the Appendix. The proposed approach significantly reduces the group sufficiency gap of in most subgroups.  The similar trend is also observed in Toxic dataset, as shown in Fig.~\ref{fig:suff_toxic}, where the proposed approach has the nearly identical and small group sufficiency gap for each race. In contrast, the baselines exhibit significant group sufficiency bias on the Asian and Latino \& other races.   

\begin{wrapfigure}{r}{7cm}
    \vspace{-12pt}
    \centering
    \includegraphics[width=70mm]{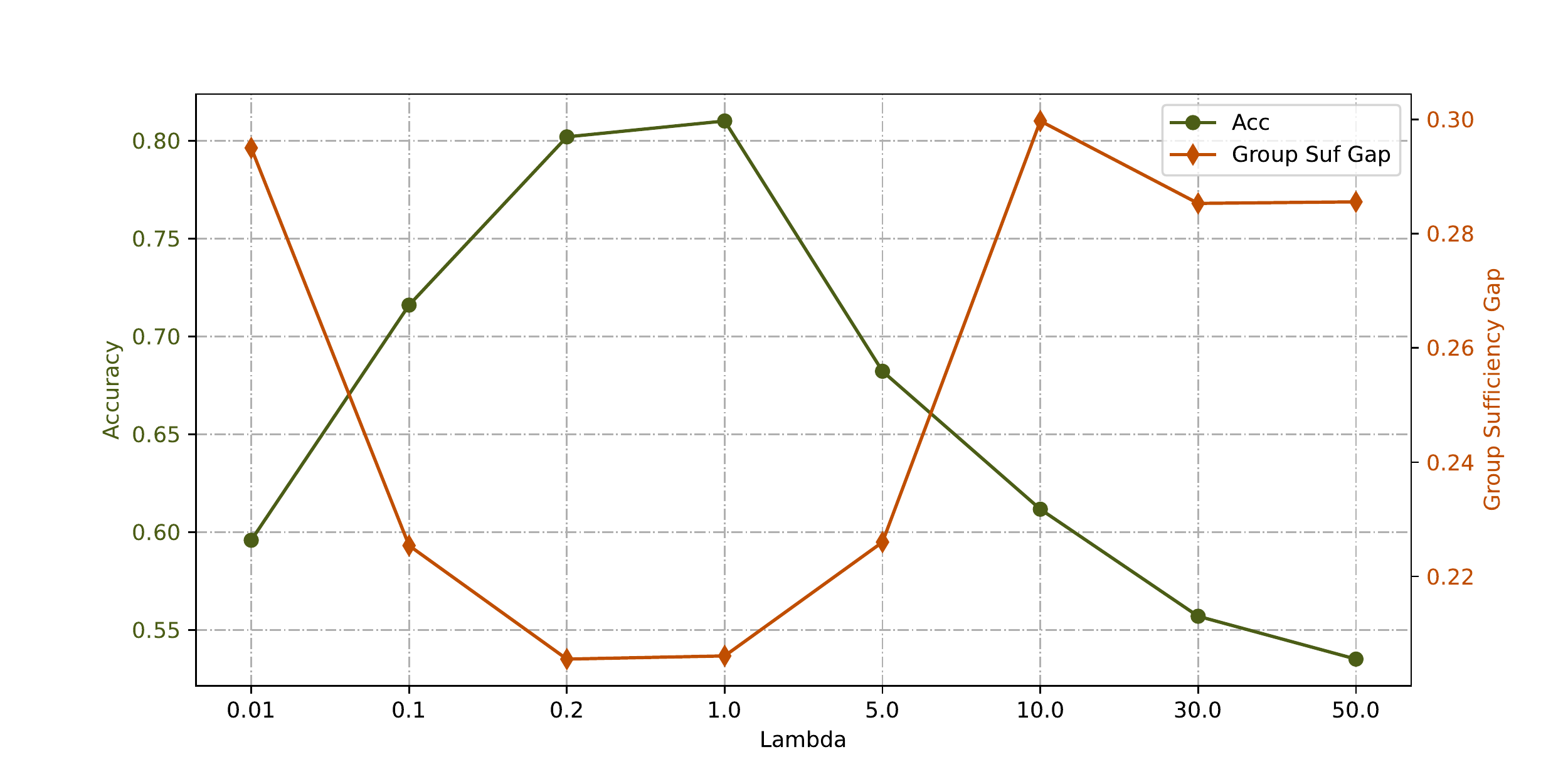}
    \caption{Analysis. Accuracy-$\textbf{Suf}_{f}$ curve under different $\lambda$ in Amazon review dataset.}
    \label{fig:amazon_lambda}
\end{wrapfigure}
\textbf{Probability Calibration} A related concept to group sufficiency is the probability calibration \citep{guo2017calibration}, which is defined as
$\E[Y|f(X)]=f(X)$ in the binary classification. We visualize the probability sufficiency of Amazon review in Fig.~\ref{fig:cal_amazon} and Toxic comments in Fig.~\ref{fig:cal_toxic}. The results suggest that the proposed approach demonstrates a consistently better probability calibration on the whole data. We then visualize the probability calibration for each subgroup, as shown in the Appendix, where the results reveal the improved probability calibration for each subgroup.

\textbf{Other sensitive attributes in Toxic comments} Apart from adopting race as sensitive attribute, we also consider other possible sensitive attributes such as gender and religion, and the results are showed in the Appendix. The results in other sensitive attributes are similar to race, with improved fairness and no loss on accuracy.

\textbf{Influence of $\lambda$. Fairness and accuracy can be simultaneously achieved.}
Theorem \ref{thm:pac_bayes} suggests that there exists an optimal $\lambda$ in the generalization error bound. Then we changed the value of $\lambda$ in Amazon dataset, as shown in Fig. \ref{fig:amazon_lambda}.

When $\lambda \to  0$, the subgroup specific parameters are simply learned from the limited samples within each subgroup. Then the fair predictor could not learn a proper prior from the subgroup specific predictor with a significant generalization error. Meanwhile, the group sufficiency gap is also large, which is consistent with \cite{pmlr-v97-liu19f}: overfitting generally degrades the group sufficiency. When $\lambda$ is set between $[0.2,1]$, the generalization error is small (with a high accuracy) and group 
sufficiency gap is kept small, implying that both fairness and accuracy can be achieved. In contrast, if we set a large value for $\lambda \gg 0$, the predictor is unable to learn from the data but from the random prior $Q$. The prediction will be completely random (accuracy $= 55\%$ when $\lambda=50$). When the predictor outputs a random guess, different from demographic parity (DP) or equalized odds (EO), the group sufficiency gap is also large.  
The analysis reveals that there exists an optimal $\lambda$ for simultaneously achieving accuracy and group sufficiency.

\section{Conclusion}
We conducted a novel analysis by simultaneously learning an informative and fair classifier for multiple or even many subgroups. We derived a novel principled algorithm. We further theoretically justified the generalization error and fair guarantees of the proposed framework. The empirical results in two real-world datasets demonstrated the effectiveness in both preserving the accuracy, as well as group sufficiency.

\section*{Discussion on Limitations}
We proposed the analysis on learning group sufficiency and informative predictors, and developed a principled approach for it. Simultaneously there are several limitations to the proposed theory and algorithm. (1) In general, group sufficiency and DP/EO are incompatible. Controlling group sufficiency, for example, would cause DP/EO degradation. This would be problematic if DP/EO were preferred in practice. (2) We also assumed that the ground truth $A$-Bayes predictors would be similar across groups. However, this assumption could be violated, resulting in a highly non-trivial scenario. Thus, in order to evaluate the conditional distribution shift, we need to consider a new setting by collect sufficient data per subgroup.

\section*{Acknowledgments and Disclosure of Financial Support}
We appreciate constructive feedback from anonymous reviewers and meta-reviewers. We also would like to thank Jun Xiao for the discussion and proof-reading the manuscript. C. Shui and C. Gagn\'e acknowledge support from NSERC-Canada and the Canada CIFAR Chairs in AI. G. Xu, J. Li, C. Ling and B. Wang are supported by Natural Sciences and Engineering Research Council of Canada (NSERC), Discovery Grants program. T. Arbel is supported by International Progressive Multiple Sclerosis Alliance, the Canada Institute for Advanced Research (CIFAR) Artificial Intelligence Chairs program, the Natural Sciences and Engineering Research Council of Canada. Q. Chen is supported by China Scholarship Council. 

\bibliography{neurips_2022}
\bibliographystyle{unsrtnat}

\section*{Checklist}

The checklist follows the references.  Please
read the checklist guidelines carefully for information on how to answer these
questions.  For each question, change the default \answerTODO{} to \answerYes{},
\answerNo{}, or \answerNA{}.  You are strongly encouraged to include a {\bf
justification to your answer}, either by referencing the appropriate section of
your paper or providing a brief inline description.  For example:
\begin{itemize}
  \item Did you include the license to the code and datasets? \answerYes{See Section~\ref{gen_inst}.}
  \item Did you include the license to the code and datasets? \answerNo{The code and the data are proprietary.}
  \item Did you include the license to the code and datasets? \answerNA{}
\end{itemize}
Please do not modify the questions and only use the provided macros for your
answers.  Note that the Checklist section does not count towards the page
limit.  In your paper, please delete this instructions block and only keep the
Checklist section heading above along with the questions/answers below.

\begin{enumerate}

\item For all authors...
\begin{enumerate}
  \item Do the main claims made in the abstract and introduction accurately reflect the paper's contributions and scope?
    \answerYes{}
  \item Did you describe the limitations of your work?
    \answerYes{The proposed theoretical analysis is restricted in the randomized algorithm.}
  \item Did you discuss any potential negative societal impacts of your work?
    \answerYes{We study the algorithmic fairness. It is worth noting that ensuring group sufficiency may degrade other fair criteria such as DP/EO.}
  \item Have you read the ethics review guidelines and ensured that your paper conforms to them?
    \answerYes{}
\end{enumerate}

\item If you are including theoretical results...
\begin{enumerate}
  \item Did you state the full set of assumptions of all theoretical results?
    \answerYes{In the paper and Appendix.}
        \item Did you include complete proofs of all theoretical results?
    \answerYes{In Appendix.}
\end{enumerate}

\item If you ran experiments...
\begin{enumerate}
  \item Did you include the code, data, and instructions needed to reproduce the main experimental results (either in the supplemental material or as a URL)?
    \answerYes{We provided the code for the reproduction.}
  \item Did you specify all the training details (e.g., data splits, hyperparameters, how they were chosen)?
    \answerYes{See the source code.}
        \item Did you report error bars (e.g., with respect to the random seed after running experiments multiple times)?
    \answerYes{We visualize the Boxplot of the results.}
        \item Did you include the total amount of compute and the type of resources used (e.g., type of GPUs, internal cluster, or cloud provider)?
    \answerNA{}
\end{enumerate}

\item If you are using existing assets (e.g., code, data, models) or curating/releasing new assets...
\begin{enumerate}
  \item If your work uses existing assets, did you cite the creators?
    \answerNA{}
  \item Did you mention the license of the assets?
    \answerNA{}
  \item Did you include any new assets either in the supplemental material or as a URL?
    \answerNA{}
  \item Did you discuss whether and how consent was obtained from people whose data you're using/curating?
    \answerNA{}
  \item Did you discuss whether the data you are using/curating contains personally identifiable information or offensive content?
    \answerNA{}
\end{enumerate}

\item If you used crowdsourcing or conducted research with human subjects...
\begin{enumerate}
  \item Did you include the full text of instructions given to participants and screenshots, if applicable?
    \answerNA{}
  \item Did you describe any potential participant risks, with links to Institutional Review Board (IRB) approvals, if applicable?
    \answerNA{}
  \item Did you include the estimated hourly wage paid to participants and the total amount spent on participant compensation?
    \answerNA{}
\end{enumerate}
\end{enumerate}

\newpage
\appendix

\section{Group sufficiency vs. demographic parity (DP) and equalized odds (EO)}\label{sec:compare}
For better understanding the properties of these three metrics, we consider the following scenario. 

Consider the score predictor $f(X)$ uniformly outputs a value in $[0,1]$ for any $x\in\calX$. i.e, $f(x)\sim \text{Unif}[0,1]$. Then it is easy to verify the demographic party and equalized odds both satisfy. Since the output of predictor is completely independent with the data. Thus we have
\[
\E[f(X)]=\E[f(X)|A],  \E[f(X)|Y]=\E[f(X)|A,Y]
\]

In contrast, the group sufficiency does not necessarily hold. Specifically we have
$\E[Y|f(X)]=\E[Y], \E[Y|f(X),A]=\E[Y|A]$, then we have
\[
\textbf{Suf}_f = \E_{A} |\E[Y]- \E[Y|A]|>0,
\]
if $A \not\!\perp\!\!\!\perp Y$.

The aforementioned counterexample suggests that group sufficiency shows different behaviors, where the pure random prediction could trivially achieve the EO/DP. 

Moreover, based on \cite{barocas-hardt-narayanan}, if $\D(X,Y,A)>0$ and $A \not\!\perp\!\!\!\perp Y$, the group sufficiency and demographic parity/equalized odds do not both hold. The aforementioned example also verifies this fact.

\section{Additional Facts of $A$-Group Bayes predictor}

For better understanding $A$-Group Bayes predictor, we could derive the following facts.

\begin{proposition}\label{propos_a_bayes}
The $A$-group Bayes predictor $f_{A}^{\text{Bayes}}(X)$:
(1) Given a subgroup $A=a$, satisfies the group sufficiency $\E[Y|f_{A=a}^{\text{Bayes}}(X)] = \E[Y|f_{A=a}^{\text{Bayes}}(X),A=a]$;
(2) is the optimal predictor under subgroup $A=a$ the binary cross-entropy loss function. \footnote{The binary cross-entropy is chosen as the prediction loss because it is widely used in classification. In fact, $f_{A}^{\text{Bayes}}$ is also the optimal predictor under square loss for each subgroup $A=a$. The entire analysis can be extended to the regression with square loss.} 
Namely, $f_{A=a}^{\text{Bayes}}(x) = \text{argmin}_{f} \calL^{\text{BCE}}_{a}(f)$, where $\calL^{\text{BCE}}_{a}(f) = \E_{(x,y)\sim\D(x,y|A=a)} -[y\log(f(x)) + (1-y)\log(1-f(x))]$.
\end{proposition}

Proposition \ref{propos_a_bayes} shows that given a subgroup $A=a$, $A$-group Bayes predictor simultaneously meets both fairness (group sufficiency) and informative (optimal predictor under binary cross-entropy loss). Unfortunately, $f^{\text{Bayes}}$ is impossible to estimate since it is related to the \emph{underlying data distribution} $\D$, which is infeasible. 
Nevertheless, we can adopt $A$-group Bayes predictor $f^{\text{Bayes}}$ to derive an upper bound of group sufficiency gap $\textbf{Suf}_f$. The upper bound then holds for any predictor $f$ that can be learned from the observed data.

\subsection{Proof of Fact 1}
We first introduce the generalized Tower rule of the conditional expectation. 
\paragraph{Generalized tower rule of conditional expectation} Let $(\Omega,\mathcal{F},P)$ be the probability space and two sub $\sigma$-algebras $\mathcal{G}_1\subseteq \mathcal{G}_2 \subseteq \mathcal{F}$ are defined. Then we have
\[
\E[\E[X|\mathcal{G}_2]|\mathcal{G}_1] = \E[X|\mathcal{G}_1]
\]

\begin{proof}
Based on the generalized tower rule, we have
\begin{align*}
    \E[Y|f_{A}^{\text{Bayes}}(X),A] & = \E[\E[Y|X,A,f_{A}^{\text{Bayes}}(X)]|f_{A}^{\text{Bayes}}(X),A] \\
    & = \E[\E[Y|X,A]|f_{A}^{\text{Bayes}}(X),A] = \E[f_{A}^{\text{Bayes}}|f_{A}^{\text{Bayes}}(X),A] = f_{A}^{\text{Bayes}}(X)
\end{align*}
Also we have:
\begin{align*}
     \E[Y|f_{A}^{\text{Bayes}}(X)] & = \E[\E[Y|f_{A}^{\text{Bayes}}(X),A,X]|f_{A}^{\text{Bayes}}(X)] \\
    & = \E[\E[Y|f_{A}^{\text{Bayes}}(X),A,X]|f_{A}^{\text{Bayes}}(X)] \\
    & = \E[\E[Y|A,X]|f_{A}^{\text{Bayes}}(X)] = \E[f_{A}^{\text{Bayes}}(X)|f_{A}^{\text{Bayes}}(X)] = f_{A}^{\text{Bayes}}(X)
\end{align*}
Combining these two equations, we have the Fact 1:
\[
\E[Y|f_{A}^{\text{Bayes}}(X),A] = \E[Y|f_{A}^{\text{Bayes}}(X)]
\]
The aforementioned proof adopts the fact $\E[Y|f_{A}^{\text{Bayes}}(X),A,X] = \E[Y|A,X]$, since the conditional expectation is uninfluenced by the $A$-group Bayes predictor, given $A,X$. 
\end{proof}

\subsection{Proof of Fact 2}
Following \cite{bishop:2006:PRML}, we can compute the optimal predictor of attribute $A=a$ under the binary cross-entropy by taking the functional derivative w.r.t. $f$:
\[
\frac{d \calL_a^{\text{BCE}}(f)}{d f}=0
\]
We have 
\begin{align*}
    & \E_{\D(x,y|a)}[f(x)-y] = 0 \\
  \to & \E_{\D(x|a)} [f(x)\E_{\D(y|x,a)} - \E_{\D(y|x,a)} y ] = 0 \\
  \to & \E_{\D(x|a)} [f(x) - \E_{\D(y|x,a)} y ] = 0
\end{align*}
Therefore, we have the optimal predictor $f^{\star}= \E_{\D(y|x,a)} y = \E[Y=y|X=a,A=a]$, the $A$-group Bayes predictor. \cite{bishop:2006:PRML} further demonstrated the optimal is unique under the binary cross entropy loss. 

\section{Upper bound of group sufficiency gap}
Before deriving the theory, we need the following lemma.
\begin{lemma}
For any predictor $f$, we have
\begin{align*}
    & \E[Y|f(X),A]= \E[f_{A}^{\text{Bayes}}(X)|f(X),A]
\end{align*}
\end{lemma}
\begin{proof}
\begin{align*}
\E[Y|f(X),A] &= \E[\E[Y|f(X),A,X]|f(X),A] \\
&= \E[\E[Y|A,X]|f(X),A] \\
&= \E[f_{A}^{\text{Bayes}}(X)|f(X),A]
\end{align*}
\end{proof}
Based on Lemma B.1, we can derive the main Theorem.
\begin{proof}
For the simplicity, we denote $\E[Y|f,A] = \E[Y|f(X),A]$ and $\E[Y|f_{A}^{\text{Bayes}},A] = \E[Y|f_{A}^{\text{Bayes}}(X),A]$. We first bound $\E_{A,X}[|\E[Y|f,A]-\E[Y|f_{A}^{\text{Bayes}},A]|]$ 
\begin{align*}
     \E_{A,X}[|\E[Y|f,A]-\E[Y|f_{A}^{\text{Bayes}},A]|] & = \E_{A,X} [|\E[f_{A}^{\text{Bayes}}|f,A]-f_{A}^{\text{Bayes}}|]\\
    & = \E_{A,X}[|\E[f_{A}^{\text{Bayes}}-f|f,A]+f-f_{A}^{\text{Bayes}}|]\\
    & \leq \E_{A,X}[|\E[f_{A}^{\text{Bayes}}-f|f,A]| + |f-f_{A}^{\text{Bayes}}|] \\
    & = 2 \E_{A,X} [|f-f_{A}^{\text{Bayes}}|]
\end{align*}
Where $\E_{A,X}[|\E[f_{A}^{\text{Bayes}}-f|f,A]|] = \E_{A,X}[\E|f_{A}^{\text{Bayes}}-f|f,A,X|] = \E_{A,X}[|\E[f_{A}^{\text{Bayes}}-f|A,X]|] =  \E_{A,X}[|f_{A}^{\text{Bayes}}-f|]$ is derived from the tower rule of conditional expectation.

Analogously, we can bound
\[
    \E_{A,X}[|\E[Y|f]-\E[Y|f_{A}^{\text{Bayes}}]|] \leq 2\E_{A,X}[|f-f_{A}^{\text{Bayes}}|]
\]

Thus the group sufficiency gap is upper bounded by:
\begin{align*}
    \textbf{Suf}_f & = \E_{A,X} [|\E[Y|f]-\E[Y|f,A]|]\\
    & = \E_{A,X} [|\E[Y|f]-\E[Y|f,A]-\E[Y|f_{A}^{\text{Bayes}}]-\E[Y|f_{A}^{\text{Bayes}},A]|]~~(\text{Using Fact 1})\\
    & \leq \E_{A,X} [|\E[Y|f]-\E[Y|f_{A}^{\text{Bayes}}]|+|\E[Y|f,A]-\E[Y|f_{A}^{\text{Bayes}},A]|]\\
    & \leq 4\E_{A,X}[|f-f_{A}^{\text{Bayes}}|]
\end{align*}
Therefore, if $A$ takes only finite value ($|\calA|<+\infty$) and follows uniform distribution with $\D(A=a)=1/|\calA|$, then we have: 
\[
\textbf{Suf}_f \leq \frac{4}{|\calA|}\sum_{a}\E_{X}[|f-f_{A=a}^{\text{Bayes}}||A=a]
\]
\end{proof}

\section{Upper bound of group sufficiency gap in randomized algorithm}
According to the definition, we have:
\begin{align*}
    \textbf{Suf}_{f} & \leq \frac{4}{|\calA|}\sum_{a} \E_{X}[|\E_{\tilde{f} \sim Q}\tilde{f}(x) - \E_{\tilde{f} \sim Q_a^{\star}}\tilde{f}(x)|+|\E_{\tilde{f} \sim Q_a^{\star}}\tilde{f}(x)-\E_{\tilde{f} \sim \D(y|x,a)}\tilde{f}(x)|]\\
    &\leq  \frac{4}{|A|}\sum_{a} \text{TV}(Q_a^{\star}\|Q) + \text{TV}(Q_a^{\star}\|\D(y|x,a))~~~(\text{Property of Total variation distance})\\
    & \leq \frac{2\sqrt{2}}{|A|}[\sum_{a}\underbrace{\sqrt{\text{KL}(Q_a^{\star}\|Q)}}_{\text{Estimation Error}} + \underbrace{\sqrt{\text{KL}(Q_a^{\star}\|\D(y|x,a))}}_{\text{Approximation Error}}]~~~(\text{Pinsker's inequality})
\end{align*}
The second line is derived from the property of Total variation distance. Note 
the scoring predictor ranges in $[0,1]$: $\tilde{f}(X=x) \in [0,1] $. 

The third line is derived from Pinsker's inequality. i.e, $\text{TV}(P\|Q)\leq \sqrt{\frac{1}{2}\text{KL}(P\|Q)}$. 

\section{Generalization upper bound}
\paragraph{Step 1} We first demonstrate the following Lemma, which is based on \citep{pentina2014pac,wainwright2019high}.
\begin{lemma} 
Let $f$ be a random variable taking value in $A$ and let $X_1,\dots,X_l$ be $l$ independent variables with each $X_k$ distributed to $\mu_k$ over the set $A_k$. For function $g_k:A \times A_{k} \to [a_k,b_k]$, $k=1,\dots,l$. Let $\zeta_k(f) = \E_{X_k\sim\mu_k} g_k(f,X_k)$ for any fixed value of $f$. Then for any fixed distribution $\pi$ on $A$ and any $\lambda$, $\delta>0$, the following inequality holds with high probability $1-\delta$ over the sampling $X_1,\dots,X_l$ for all distribution $\rho$ over $A$. 
\[
\E_{f\sim\rho} \sum_{k=1}^{l} \zeta_{k}(f) - \E_{f\sim\rho} \sum_{k=1}^{l} g_k(f,X_k) \leq \frac{1}{\lambda}\left(\text{KL}(\rho\|\pi)+ \frac{\lambda^2}{8}\sum_{k=1}^{l}(b_k-a_k)^2 +\log\frac{1}{\delta} \right) 
\]
\end{lemma}

\paragraph{Step 2} Then we could use the aforementioned Lemma to demonstrate the main theorem.
\begin{proof}
We adopt the lemma for the union of the whole training samples $S=\cup_{a\in\calA} S_{a}$. 

We set 
\begin{align*}
    \rho = \underbrace{(Q_1 \otimes Q_2 \otimes \dots \otimes Q_{|\calA|})}_{|\calA|~~\text{times}} \quad\quad\quad
     \pi= \underbrace{(Q \otimes Q \otimes \dots \otimes Q)}_{|\calA|~~\text{times}} 
\end{align*}

We also set
$X_k = (x_{i}^{a}, y_{i}^{a})$, $l=|\calA|m$, $f = (\tilde{f}_{1},\dots,\tilde{f}_{a},\dots,\tilde{f}_{|\calA|})$, $g_k(f,X_k) = \frac{1}{|\calA|m} \ell^{\text{BCE}}(\tilde{f}_a (x_{i}^a),y_{i}^a)$. 
Since we adopt the binary cross entropy loss, $a_k=0$ and $b_k=L/(|\calA|m)$, then with high probability $1-\delta$, we have:
\begin{align*}
\frac{1}{|\calA|}\sum_{a} \E_{\tilde{f}_a \sim Q_a} \calL_{a}^{\text{BCE}}(\tilde{f}_a) &\leq \frac{1}{|\calA|}\sum_{a} \E_{\tilde{f}_a \sim Q_a} \hat{\calL}_{a}^{\text{BCE}}(\tilde{f}_a) \\
&+ \frac{1}{\lambda}(\text{KL}(Q_1\otimes\dots\otimes Q_{|\calA|}\|Q\otimes\dots\otimes Q)+\log(\frac{1}{\delta})) + \frac{\lambda L}{8|\calA|m}
\end{align*}

Through the decomposition property of KL divergence, we finally have:
\begin{align*}
  \frac{1}{|\calA|}\sum_{a} \E_{\tilde{f} \sim Q_a} \calL_{a}^{\text{BCE}}(\tilde{f})  & \leq \frac{1}{|\calA|}\sum_{a} \E_{\tilde{f} \sim Q_a} \hat{\calL}_{a}^{\text{BCE}}(\tilde{f}) + L\sqrt{\frac{1}{2|\calA|m}(\sum_{a}\text{KL}(Q_a\|Q)+\log(\frac{1}{\delta}))}\\
    & \leq \frac{1}{|\calA|}\sum_{a} \E_{\tilde{f} \sim Q_a} \hat{\calL}_{a}^{\text{BCE}}(\tilde{f}) + \frac{L}{\sqrt{|\calA|m}}\sum_{a} \sqrt{\text{KL}(Q_a\|Q)} + L\sqrt{\frac{\log(1/\delta)}{|\calA|m}}
\end{align*}
\end{proof}

\section{Computing group sufficiency gap from the data}\label{sec:comp_g_cal}
In this paper, we need to compute the conditional expectation from the data. i.e,
\[
\E[Y|f(X)], ~~~~\E[Y|f(X),A=a],
\]
where we have observed data $\{S_a\},a\in\calA$. Since $f(x)$ is a continuous value, ranging from $[0,1]$. Then we split $[0,1]$ into sperate interevals:
\[
[0,\epsilon_1], [\epsilon_1,\epsilon_2], \dots, [\epsilon_{N},1]
\]
We compute the expectation and conditional expectation within each interval. i.e:
\[
(\E[f(X) \mathbf{1}_{\{f(X) \in [\epsilon_i,\epsilon_{i+1}]\}}] , \E[Y|f(X) \in [\epsilon_i,\epsilon_{i+1}]]) = (p_i,q_i)
\]

\[
(\E[f(X) \mathbf{1}_{\{f(X) \in [\epsilon_i,\epsilon_{i+1}],A=a\}}] , \E[Y|f(X) \in [\epsilon_{i},\epsilon_{i+1}],A=a]) = (p^a_i,q^a_i)
\]

Then for each group $A=a$, the group sufficiency gap is computed as:
\[
\textbf{Suf}_{f}(A=a) = \sum_{i} |q_i - \underbrace{(q_i^a  + \frac{p_i-p_i^a}{p^a_{i+1}-p^a_i}(q_i^{a+1}-q_i^{a}))}_{\text{Linear Interpolation}}|
\]
We use the linear interpolation if the average values in each interval are not equal. Then the group sufficiency can be formulated as:

\[
\textbf{Suf}_f = \frac{1}{|\calA|} \sum_{a} \textbf{Suf}_{f}(A=a)
\]
We assign the $\D(A=a)=\frac{1}{|\calA|}$ as uniform distribution for ensuring fairness for each subgroup.

\paragraph{Discussions}
In general:
\[
\E[Y|f(X)] \neq \frac{1}{|\calA|}\sum_{a} \E[Y|f(X),A=a] 
\]
The demonstration is straightforward. By using the Bayes rule, we have 
\[
\E[Y|f(X)] = \sum_{a} \D(A=a|f(X))\E[Y|f(X),A=a] 
\]
Thus iff $\D(A=a|f(X))=\frac{1}{|\calA|}$, we have the equivalent form. Intuitively,  $\D(A=a|f(X))$ refers the 
conditional probability of $A=a$, given the predicted score $f(X)$, which is related to the group membership inference such as \cite{hu2021membership}.  If $\D(A=a|f(X))$ is large, the subgroup index can be easily revealed via the algorithm output. If the algorithm can fully preserve the privacy, then $\D(A=a|f(X)) = \frac{1}{|\calA|}$.

\section{Experimental Details}\label{sec:add_exp_detail}
In this part, we proposed a detailed description of the dataset and experiments settings.

\subsection{Amazon review dataset}
The experiment is adapted from the protocol of \citep{koh2021wilds}. Specifically, we convert the original review score (ranging from 1-5) to the binary label: the positive review (score $\geq 4$) and negative review (score $\leq 3$). We sample and then fix 200 users from the original dataset, which contains the training (75-400 samples per user) , validation (75 samples per user), and test sets (75 samples per user). 

The total training epoch is $100$. In each training epoch, we sample a small subset of users ($N_{\text{user}}=20$), then for each user we sample $50$ samples with replacement. The early stopping strategy is also adopted. 

We adopt 4-layers fully connected neural network as the model, where the weights of the model follows the Gaussian distribution $Q_a$ or $Q$. The trade-off coefficient $\lambda$ ranges from $[0.01,50]$ and we fix $\lambda=0.4$ in the evaluation. We set all the Monte-Carlo samples as 5. More implementation details of experiments and parameter settings can be found in the code.

\subsection{Toxic comments}
We adopt the toxic comment dataset \cite{borkan2019nuanced} to predict whether the text comment is toxic or not, which has been observed the significant performance degradation on particular sub-populations. 

Following \cite{borkan2019nuanced,koh2021wilds}, we first choose the \emph{race} as sensitive attribute, which includes Black, White, Asian and Latino \& others (4 subgroups). 

The total training epoch is $100$. In each training epoch, we sample all the subgroups with the same sample size with replacement ($N=50$). The early stopping strategy is also adopted. The training ($N=33188$) validation ($N=3438$) and test set ($N=9744$) are following the protocol in \cite{koh2021wilds}.

We also consider the following sensitive attributes:
\begin{enumerate}
    \item Religion. The religion includes Christian, Jewish, Muslin and others (such as Hindu, Buddhist, atheist).
    \item Gender. The gender includes male, female and others (such as homosexual\_gay\_or\_lesbian, bisexual, transgender, other\_gender).
\end{enumerate}

Because toxic comments are marked by multiple annotators, we determine that the comment is toxic if at least half of the annotators mark it. In the implementation, we also adopt the DistilBERT \cite{sanh2019distilbert} to extract the embedding with dimension $\R^{768}$. 

We adopt 4-layers fully connected neural network as the model, where the weights of the model follows the Gaussian distribution $Q_a$ or $Q$. The trade-off coefficient $\lambda$ ranges from $[0.01,50]$ and we fix $\lambda=0.6$ in the evaluation. We set all the Monte-Carlo samples as five $N=5$. More implementation details of experiments and parameter settings can be found in the code.

\section{Additional Results in Amazon review}
\begin{figure}[h]
    \centering
    \includegraphics[width=1\textwidth]{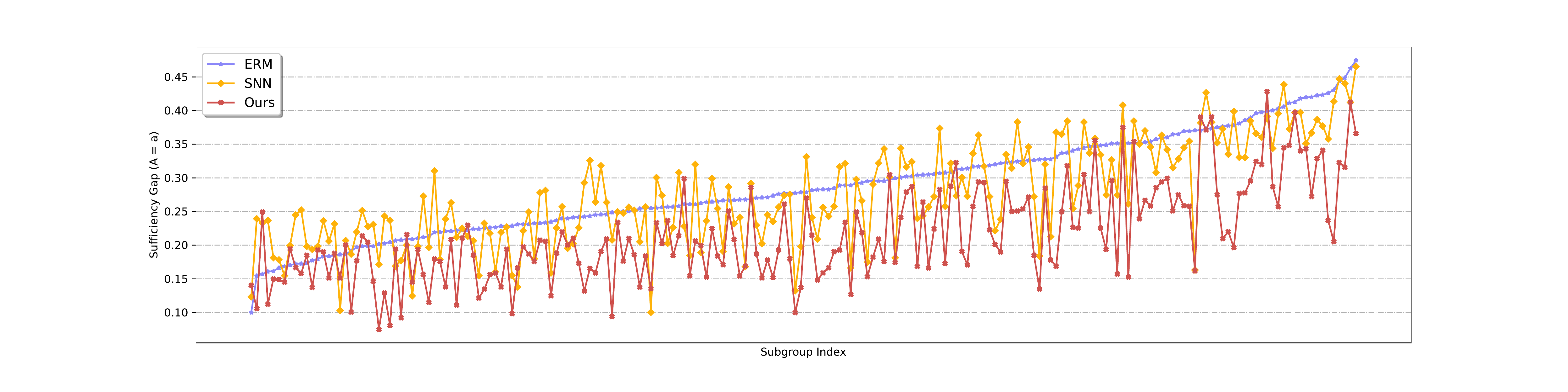}
    \caption{Group sufficiency gap on the specific subgroup $A=a$, which is the gap between $\E[Y|f(X)]$ and $\E[Y|f(X),A=a]$. We visualize the results for the entire users of Amazon Review dataset.}
    \label{fig:amazon_full_gap}
\end{figure}

\begin{figure}[h]
    \centering
    \includegraphics[width=80mm]{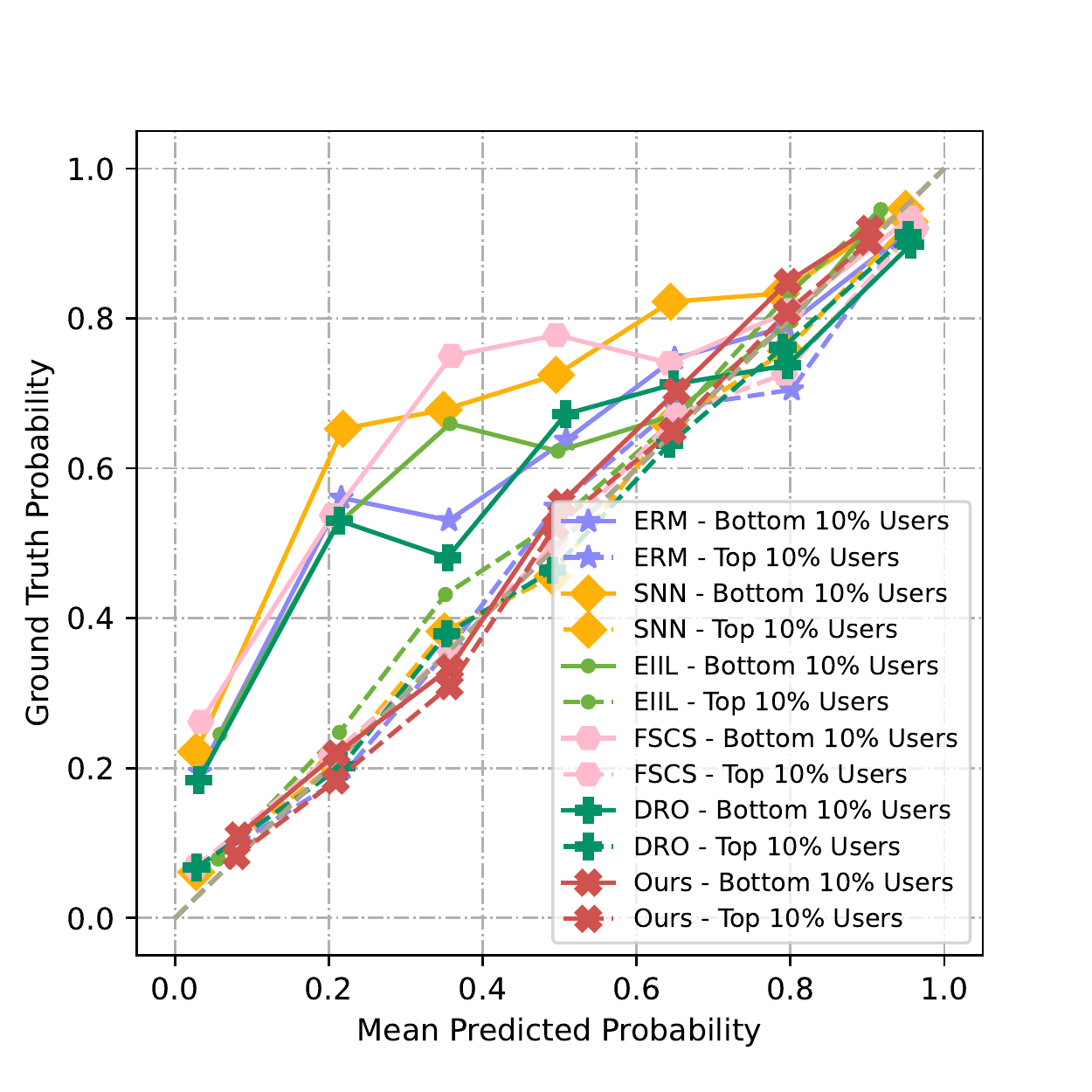}
    \caption{Probability calibration over aggregated users of Amazon Review dataset, where we sort the group sufficiency gap for each subgroup $A=a$, then we visualize the probability calibration curve for the top $10\%$ and bottom $10\%$ users. Since each user has quite limited labels, we aggregate the top $10\%$ and bottom $10\%$ and visualize the probability calibration. The results depict consistently better probability calibration of the proposed approach.}
    \label{fig:amazon_proba_cal_user}
\end{figure}

Additional results of each subgroup's fair performance and the probability calibration on the Amazon Review dataset are shown in Fig.~\ref{fig:amazon_full_gap} and Fig.~\ref{fig:amazon_proba_cal_user}.

\section{Additional Results in Toxic Comments}\label{sec:addtional_toxic}
\subsection{Race as sensitive attribute: probability calibration}
\begin{figure}[h]
\centering
  \subfigure[ERM]{\includegraphics[width=45mm]{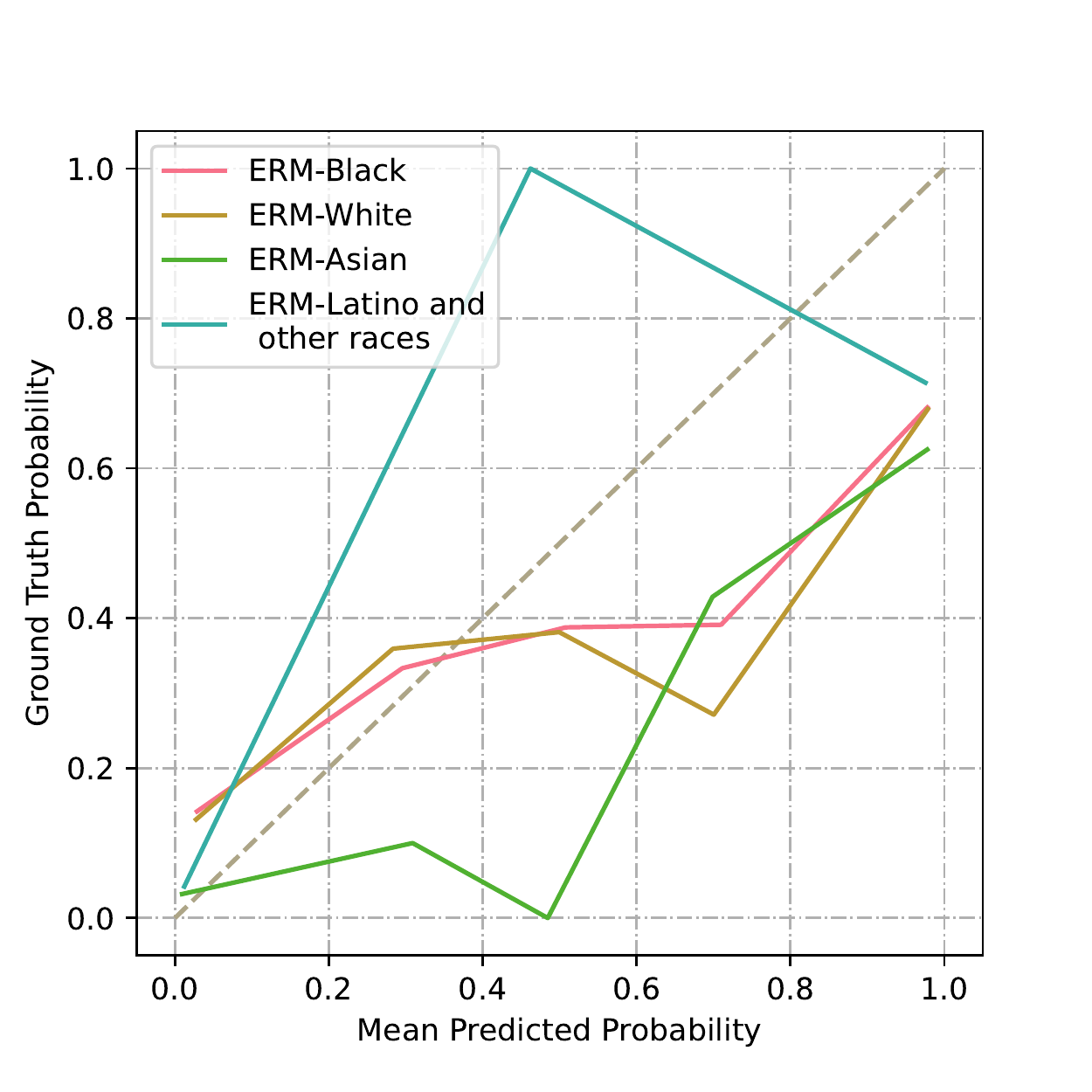}}
  \quad
  \subfigure[SNN]{\includegraphics[width=45mm]{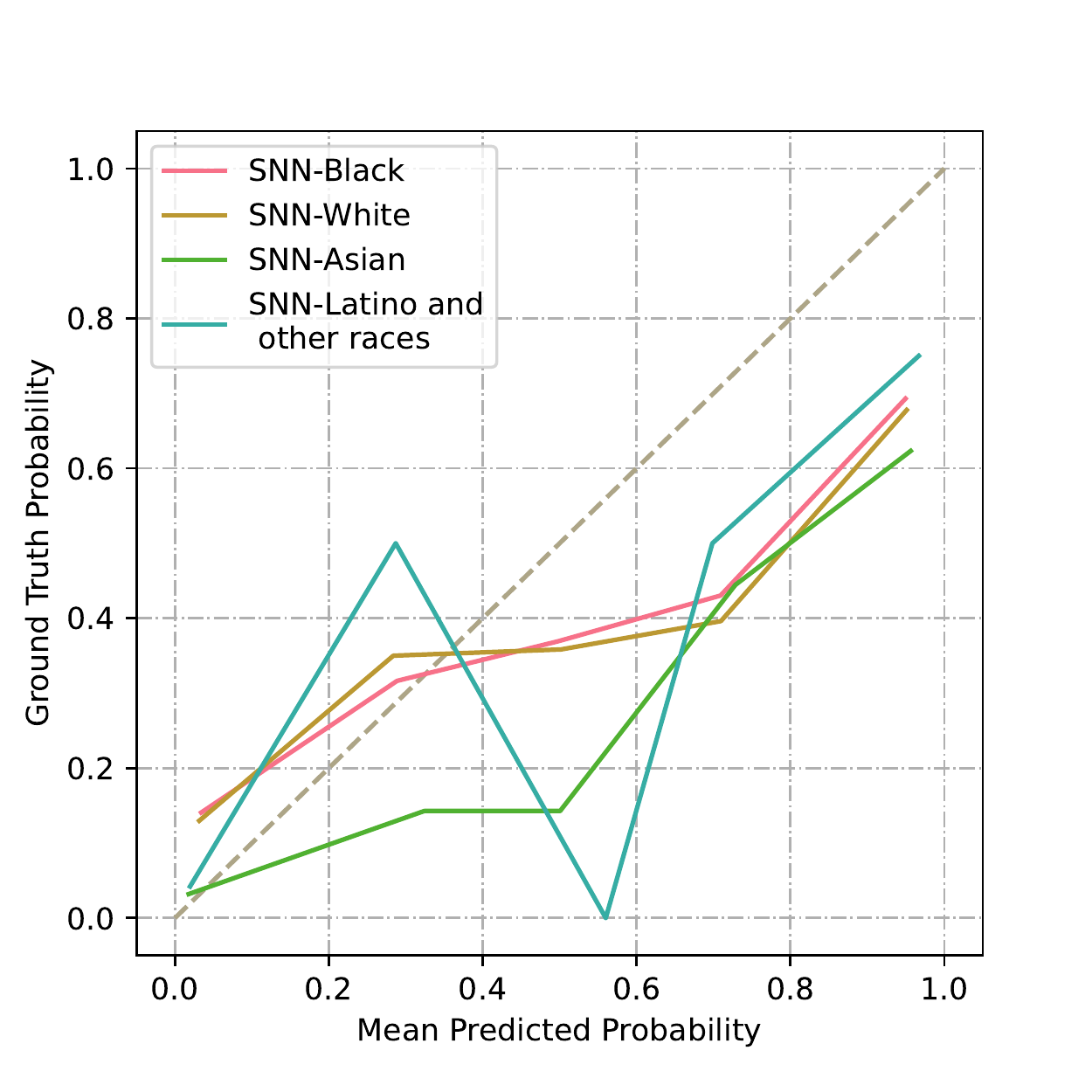}}
  \subfigure[Ours]{\includegraphics[width=45mm]{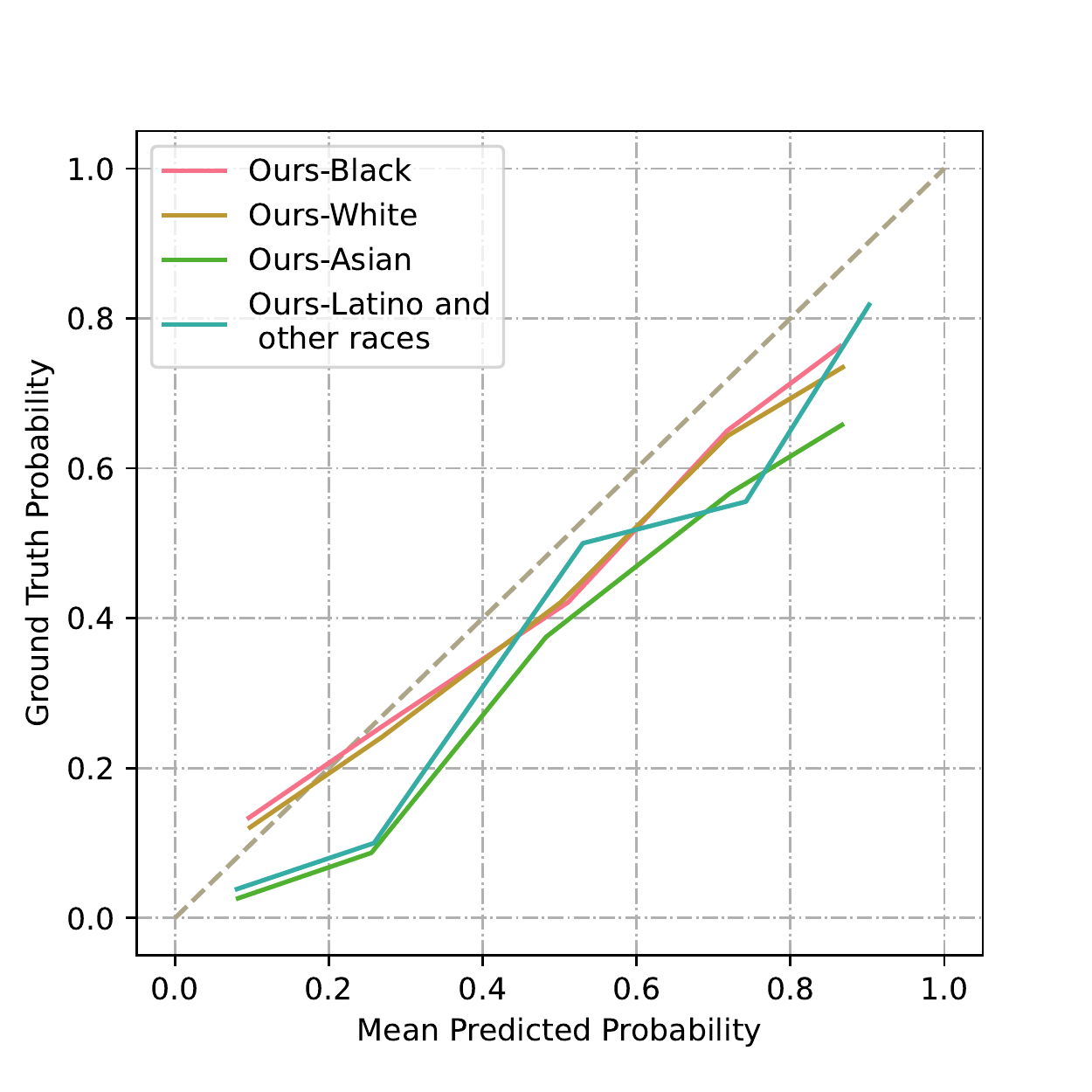}}
  \caption{Ablation. Results in Toxic dataset (Race). Visualization of probability calibration for each subgroup $A=a$, i.e. $(f(X), \E[Y|f(X),A=a])$. The proposed approach shows significant improved probability calibration for each subgroup.}\label{fig:toxic_rece_proba_cal}
\end{figure}

The additional results of the probability calibration on the Toxic Comments (Race) dataset is shown in Fig.~\ref{fig:toxic_rece_proba_cal}.

\subsection{Religion as sensitive attribute}
\begin{figure}[h]
\centering
  \subfigure[Accuracy-$\textbf{Suf}_{f}$]{\includegraphics[width=45mm]{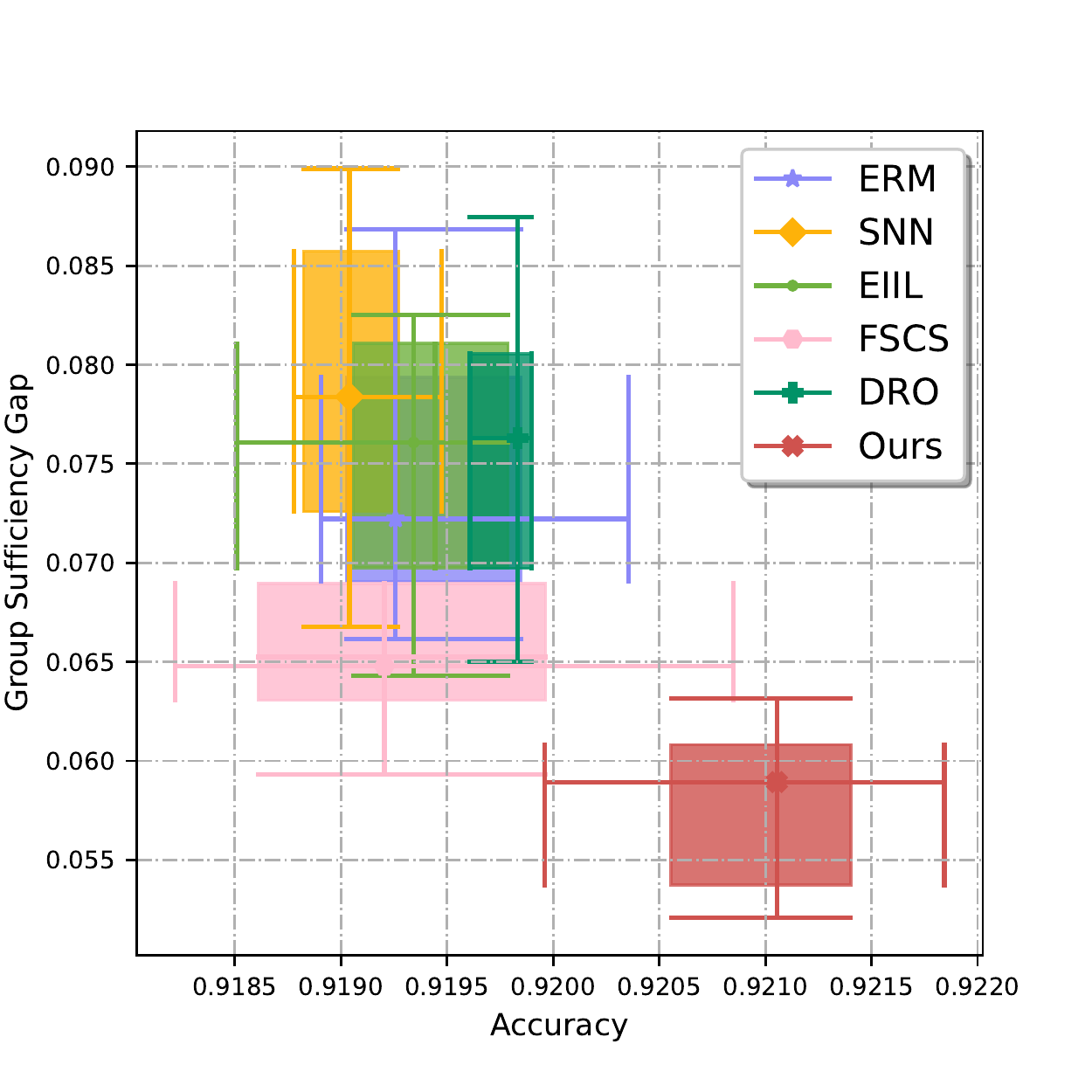}}
  \quad
  \subfigure[$\textbf{Suf}_{f}$ on each subgroup]{\includegraphics[width=45mm]{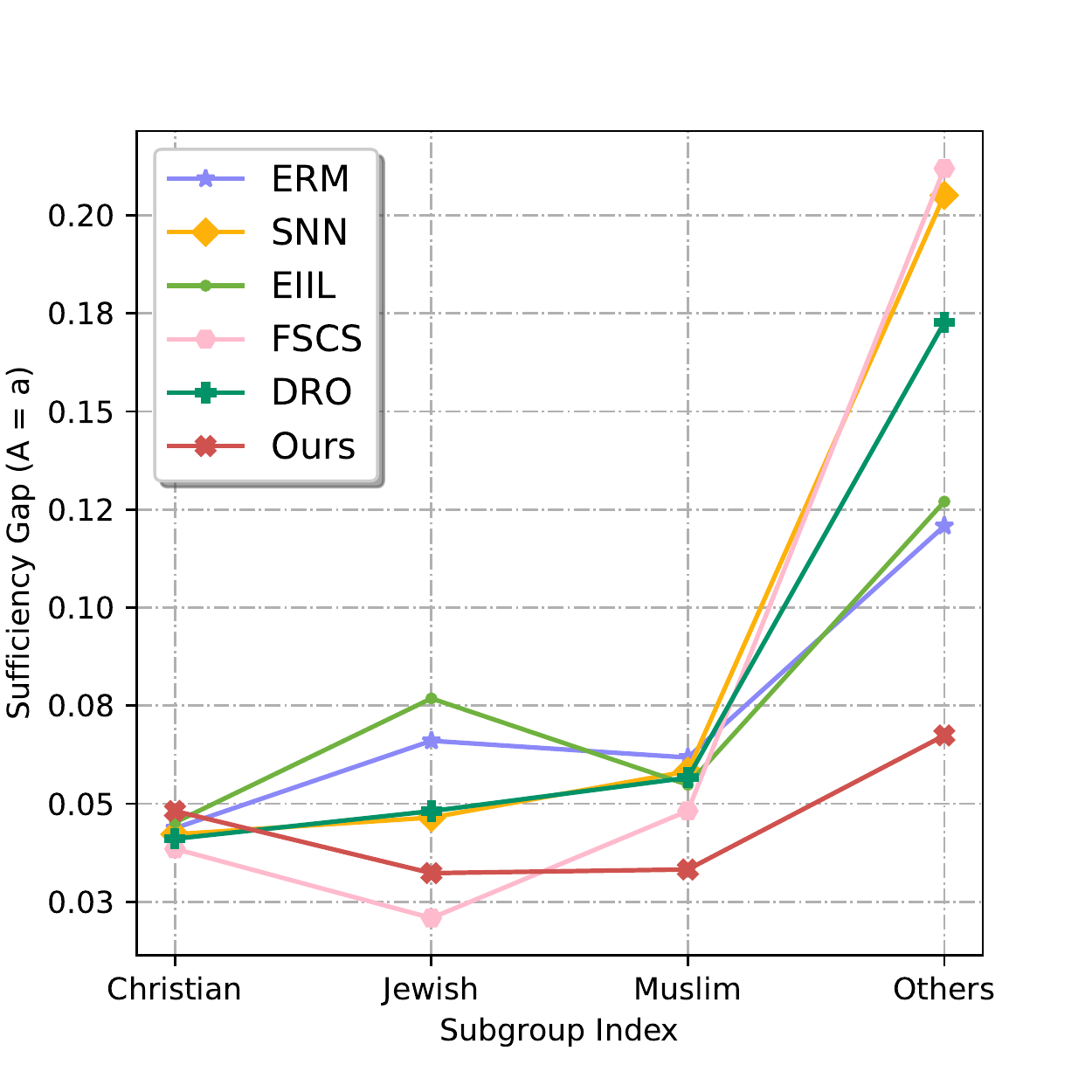}}
  \quad
  \subfigure[Probability calibration]{\includegraphics[width=45mm]{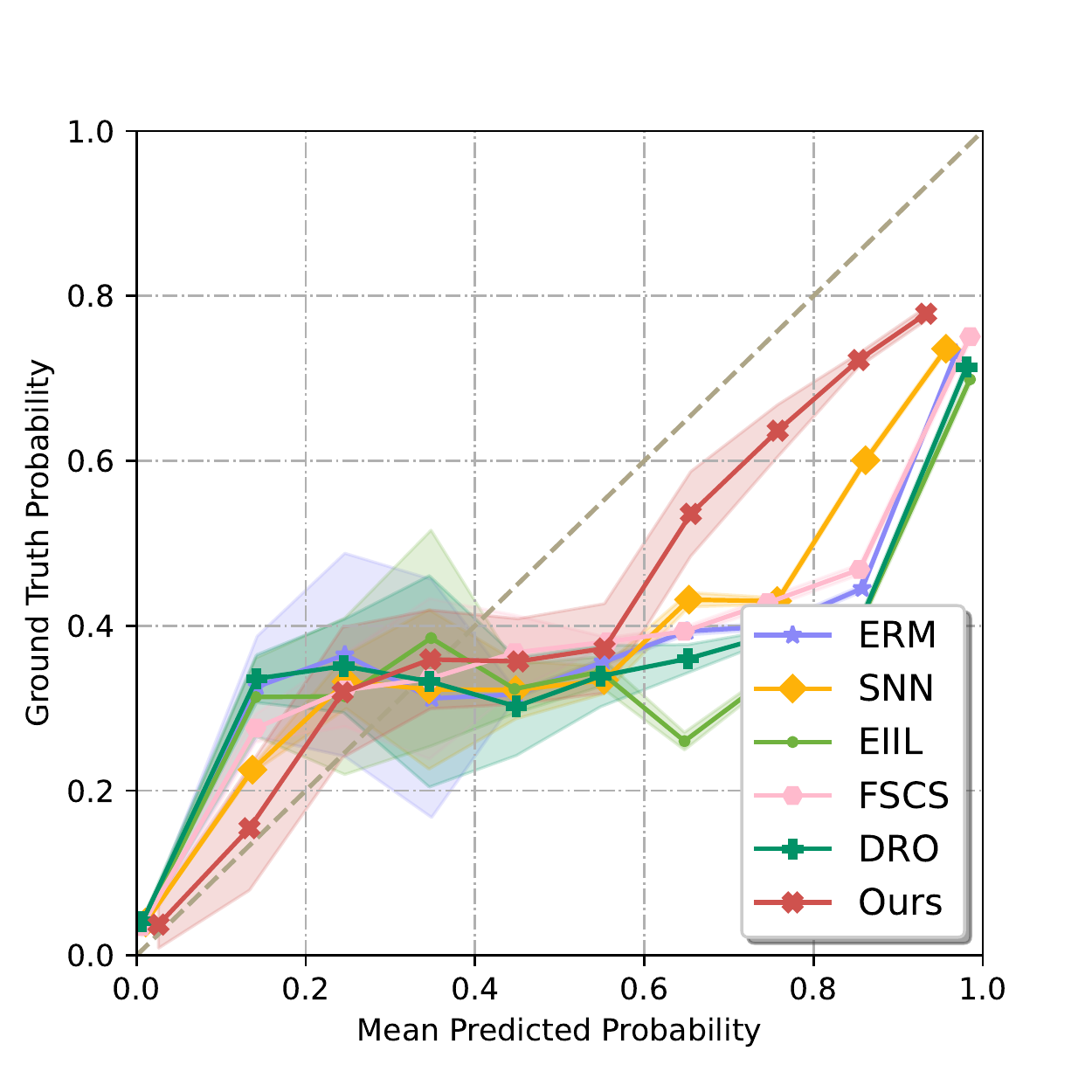}}
  \caption{Results in Toxic dataset (Religion). Boxplot of accuracy and group sufficiency gap $\textbf{Suf}_f$ with 5 repeats: median, 75th percentile and minimum-maximum value. (b)  Group sufficiency gap on the specific subgroup $A=a$, which is the gap between $\E[Y|f(X)]$ and $\E[Y|f(X),A=a]$. (c) Visualization of probability calibration over 5 repeats with mean and standard deviation. i.e $(f(X), \E[Y|f(X)])$.}\label{fig:toxic_religions}
\end{figure}

\begin{figure}[h]
\centering
  \subfigure[ERM]{\includegraphics[width=45mm]{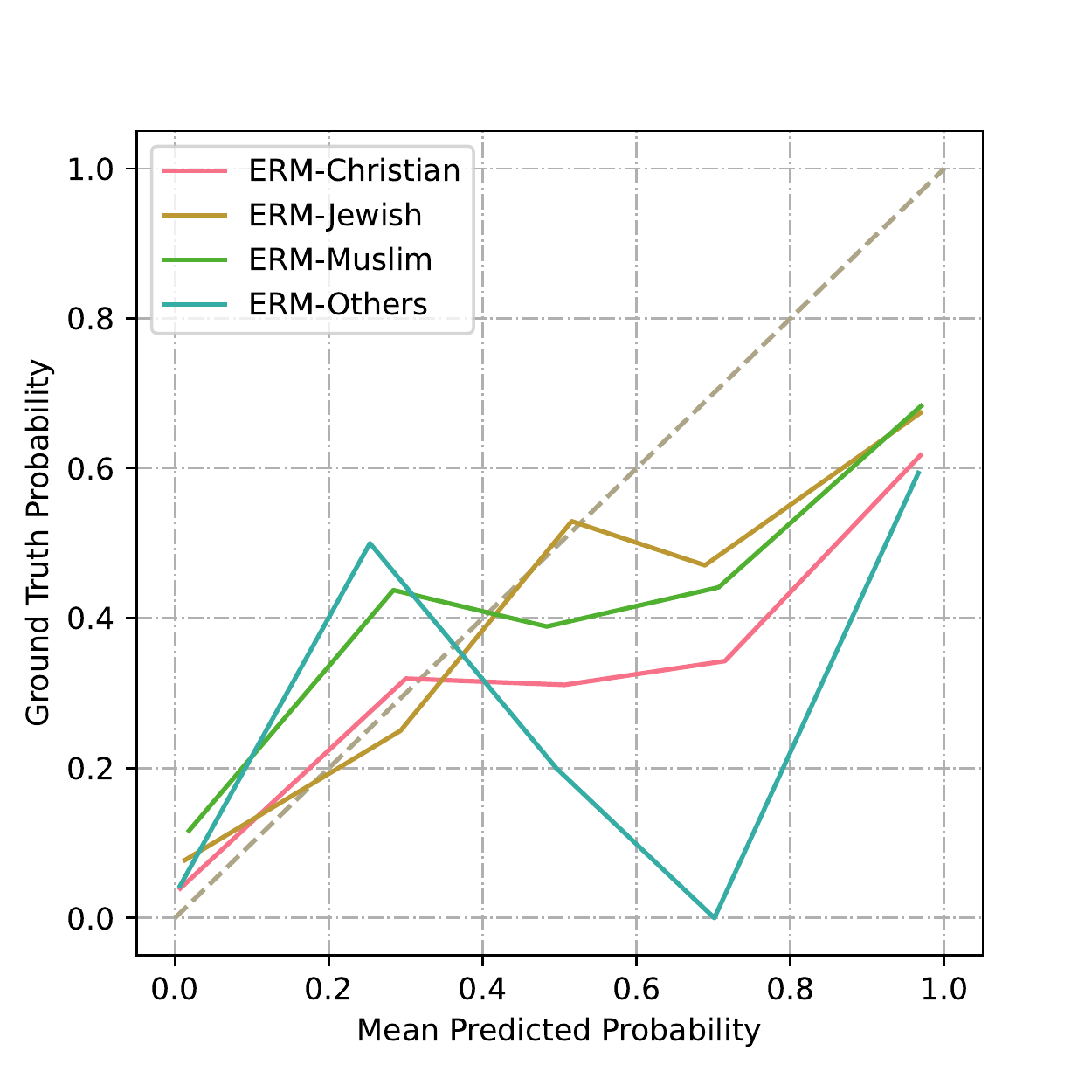}}
  \quad
  \subfigure[SNN]{\includegraphics[width=45mm]{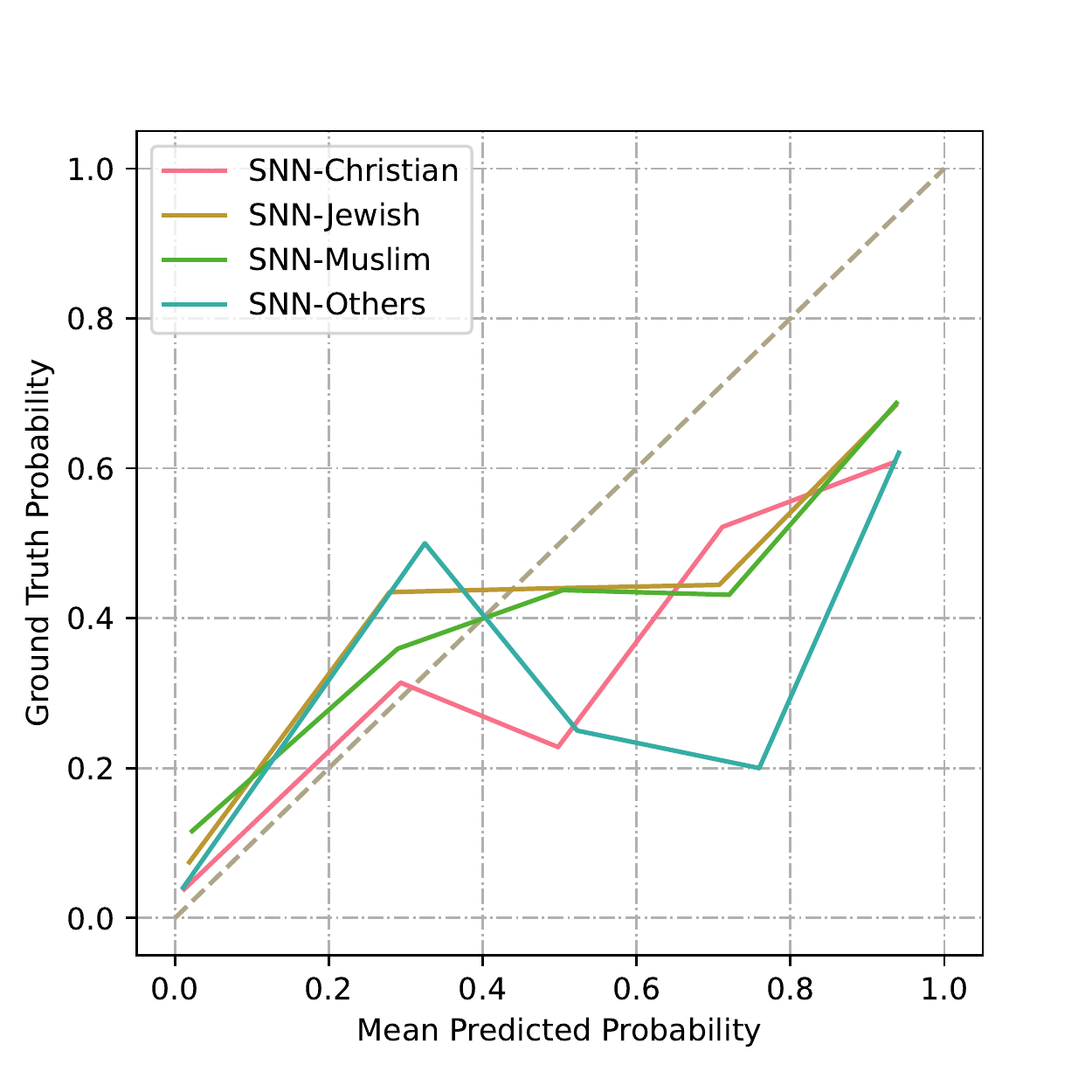}}
  \subfigure[Ours]{\includegraphics[width=45mm]{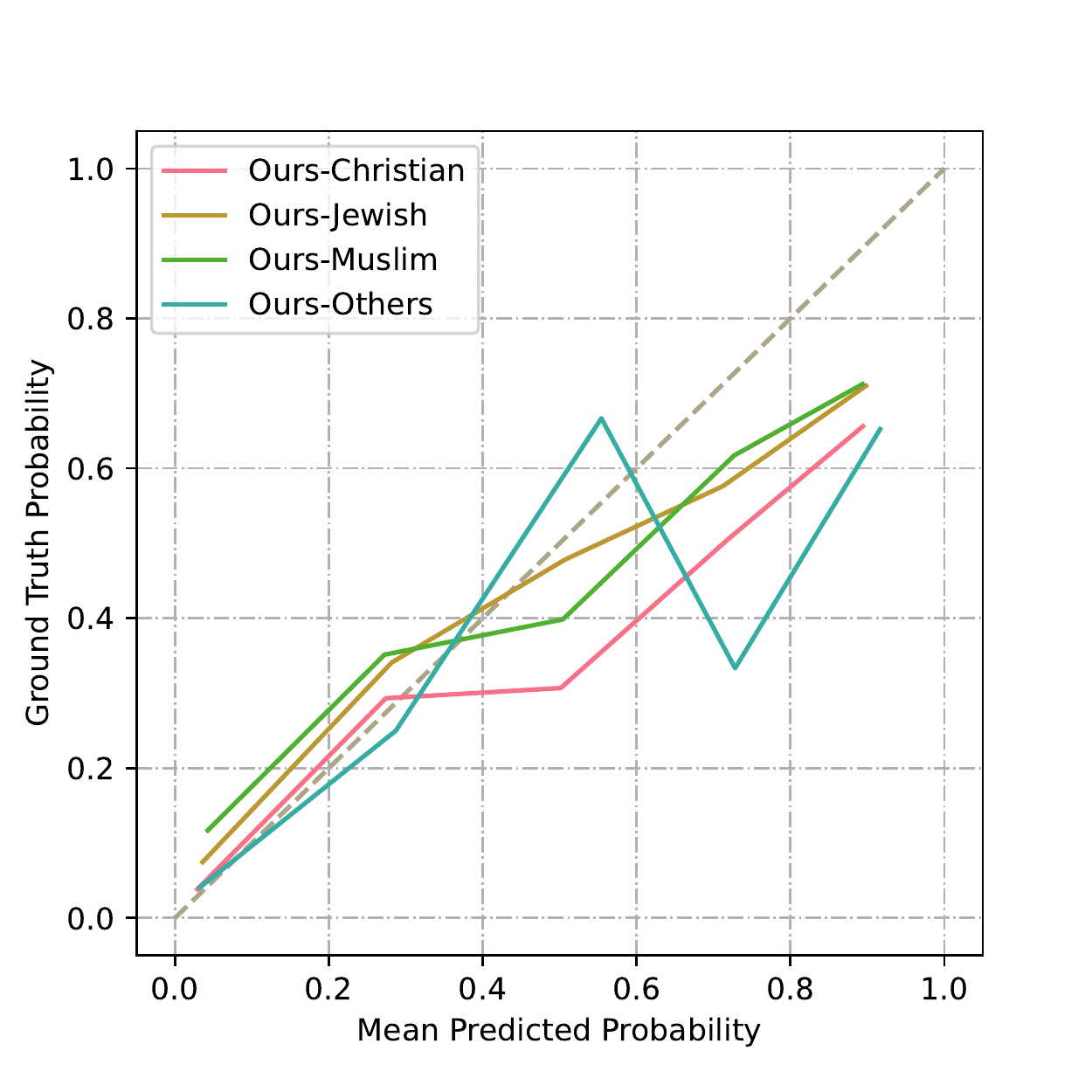}}
  \caption{Ablation. Results in Toxic dataset (Religion). Visualization of probability calibration for each subgroup $A=a$, i.e. $(f(X), \E[Y|f(X),A=a])$. The proposed approach shows significant improved probability calibration for each subgroup.}\label{fig:toxic_religion_proba_cal}
\end{figure}

The additional results of each subgroup's fair performance and the probability calibration on the Toxic Comments (Religion) dataset are shown in Fig.~\ref{fig:toxic_religions} and Fig.~\ref{fig:toxic_religion_proba_cal}.

\subsection{Gender as sensitive attribute}
\begin{figure}[h]
\centering
  \subfigure[Accuracy-$\textbf{Suf}_{f}$]{\includegraphics[width=45mm]{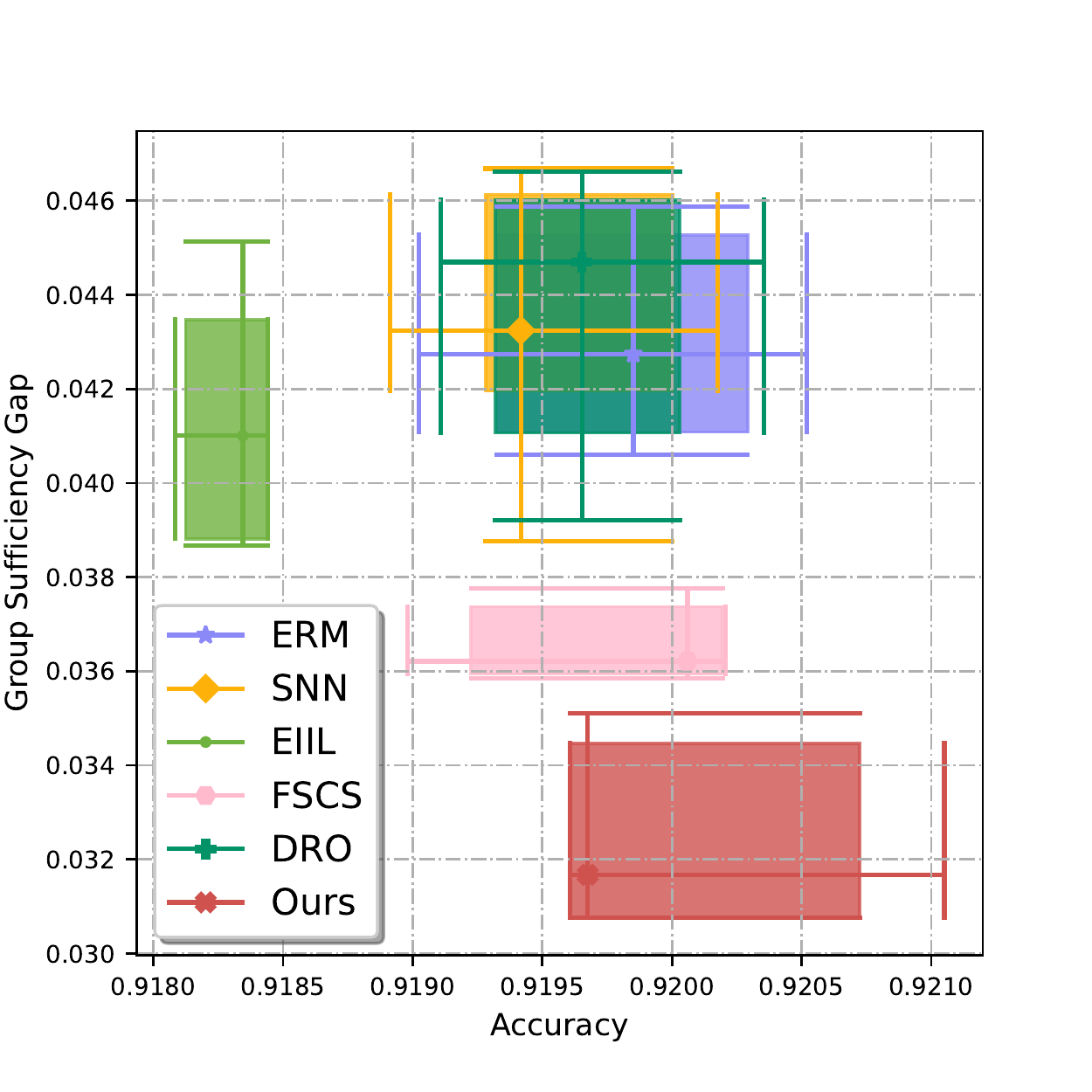}}
  \quad
  \subfigure[$\textbf{Suf}_{f}$ on each subgroup]{\includegraphics[width=45mm]{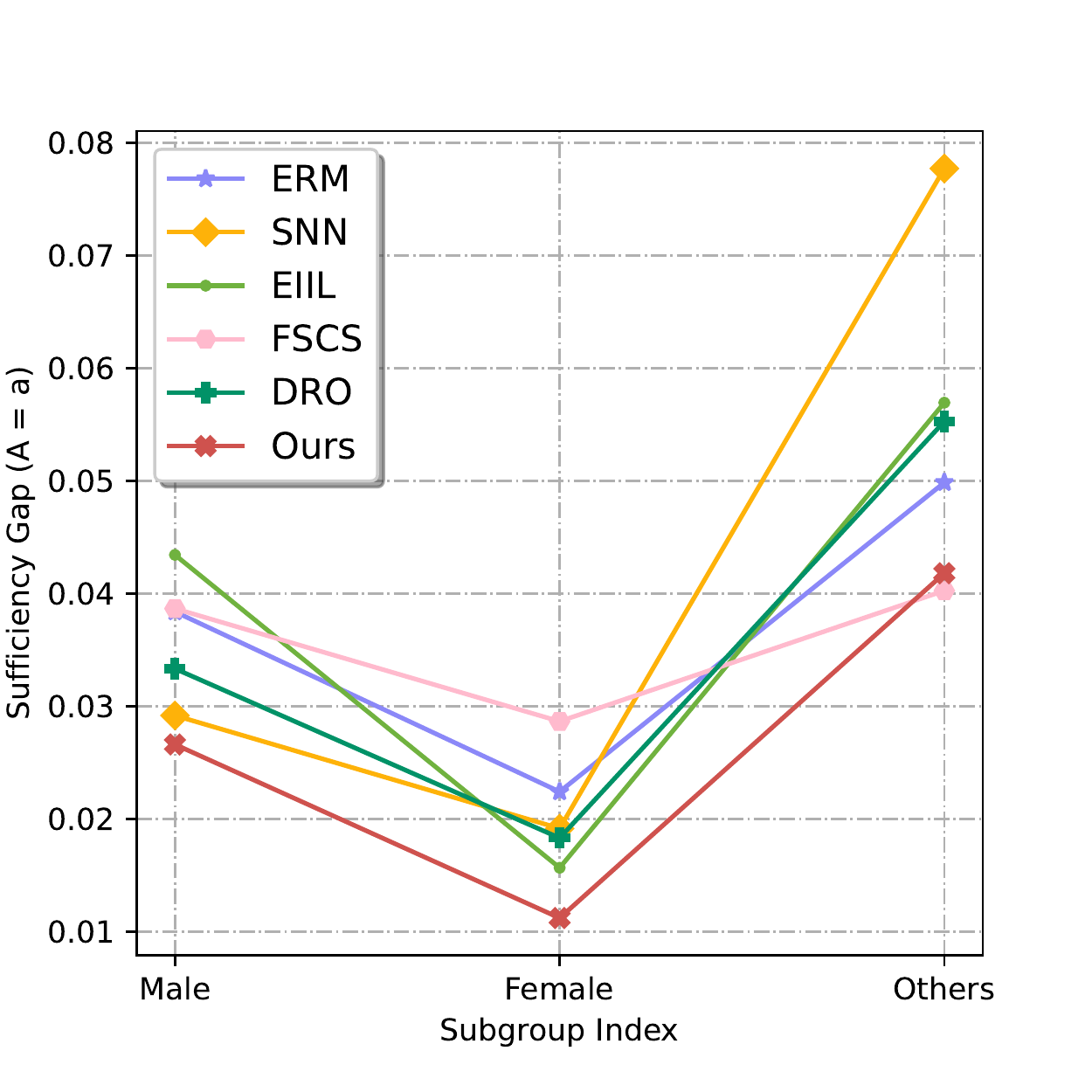}}
  \quad
  \subfigure[Probability calibration]{\includegraphics[width=45mm]{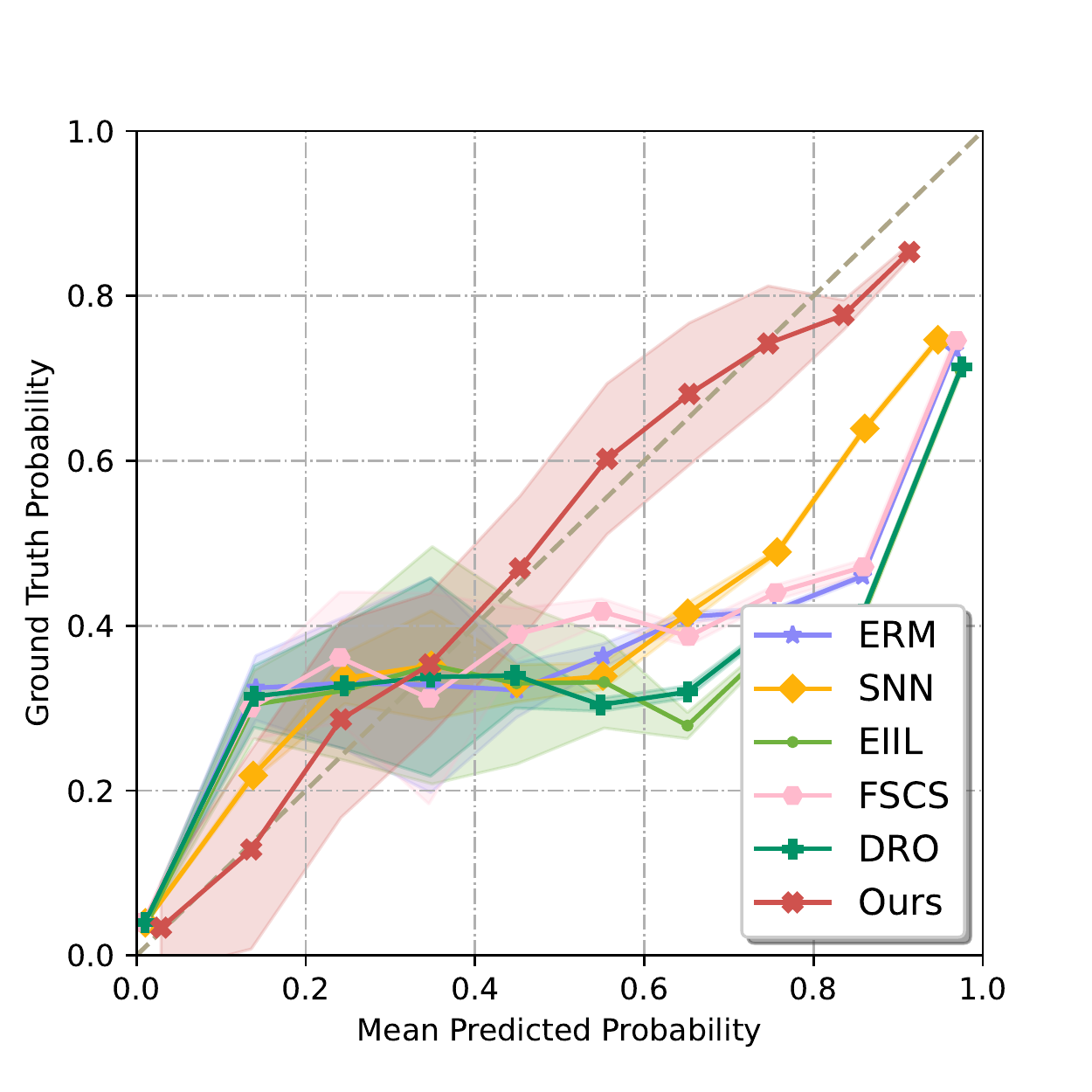}}
  \caption{Results in Toxic dataset (Gender). Boxplot of accuracy and group sufficiency gap $\textbf{Suf}_f$ with 5 repeats: median, 75th percentile and minimum-maximum value. (b)  Group sufficiency gap on the specific subgroup $A=a$, which is the gap between $\E[Y|f(X)]$ and $\E[Y|f(X),A=a]$. (c) Visualization of probability calibration over 5 repeats with mean and standard deviation. i.e $(f(X), \E[Y|f(X)])$.}\label{fig:toxic_gender}
\end{figure}

\begin{figure}[h]
\centering
  \subfigure[ERM]{\includegraphics[width=45mm]{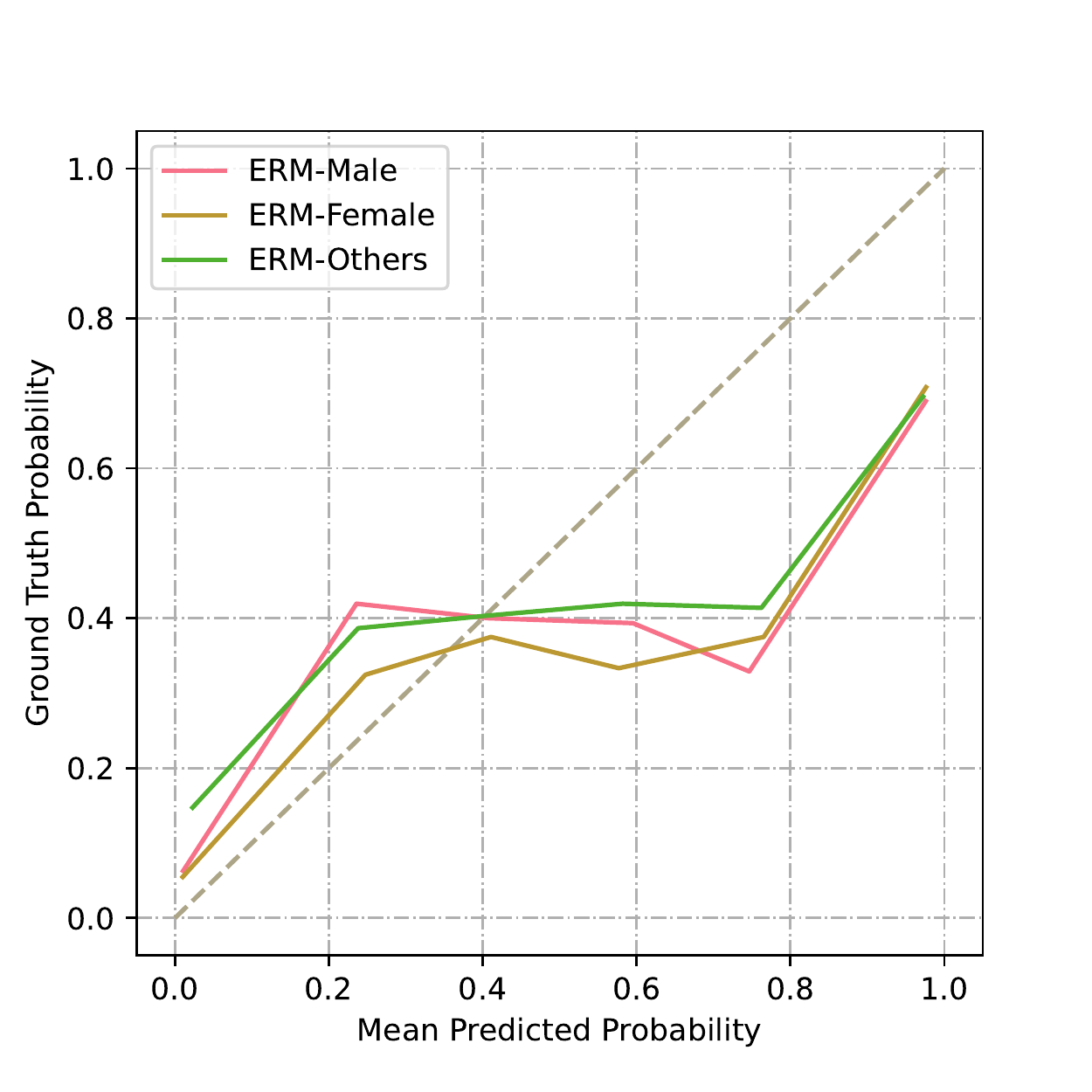}}
  \quad
  \subfigure[SNN]{\includegraphics[width=45mm]{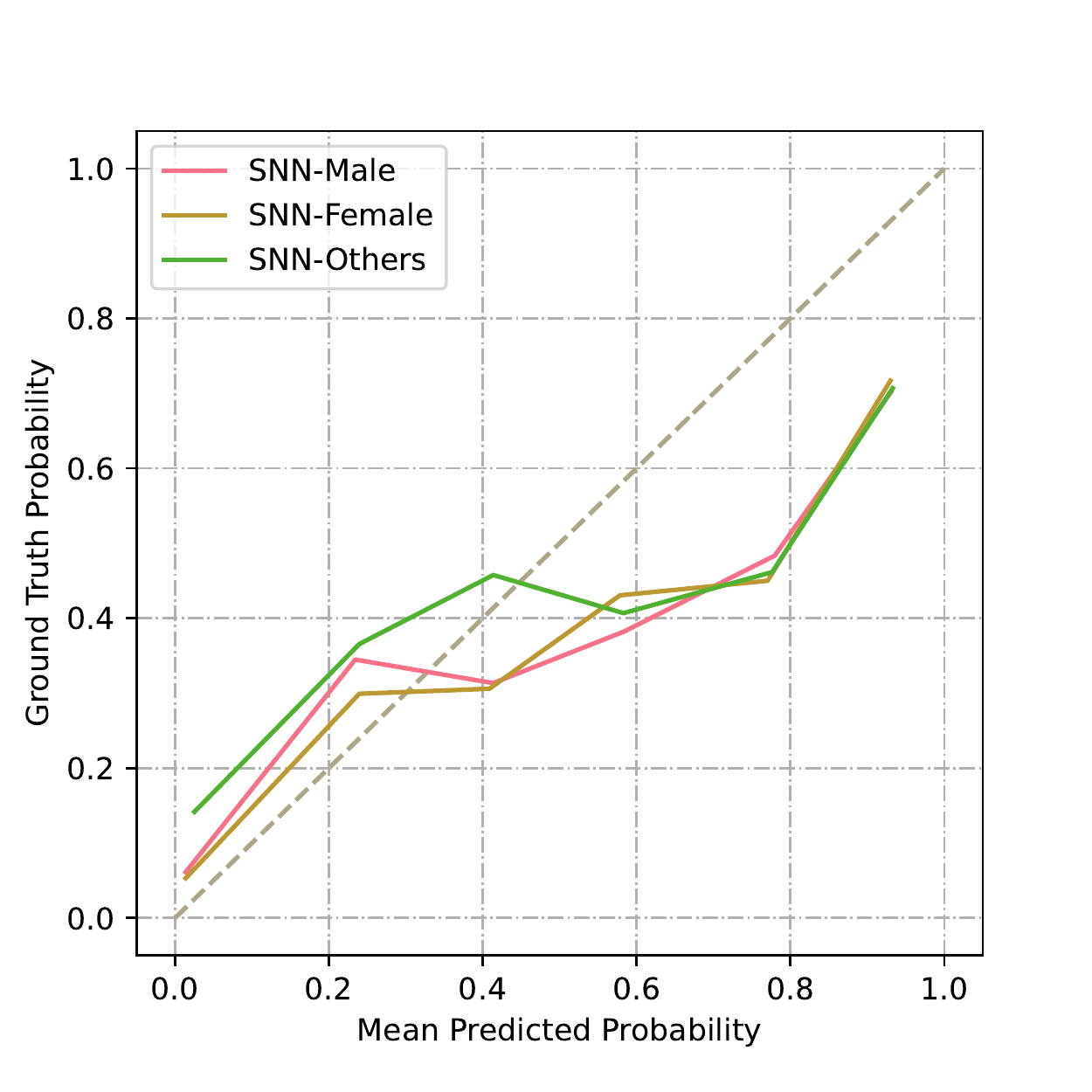}}
  \subfigure[Ours]{\includegraphics[width=45mm]{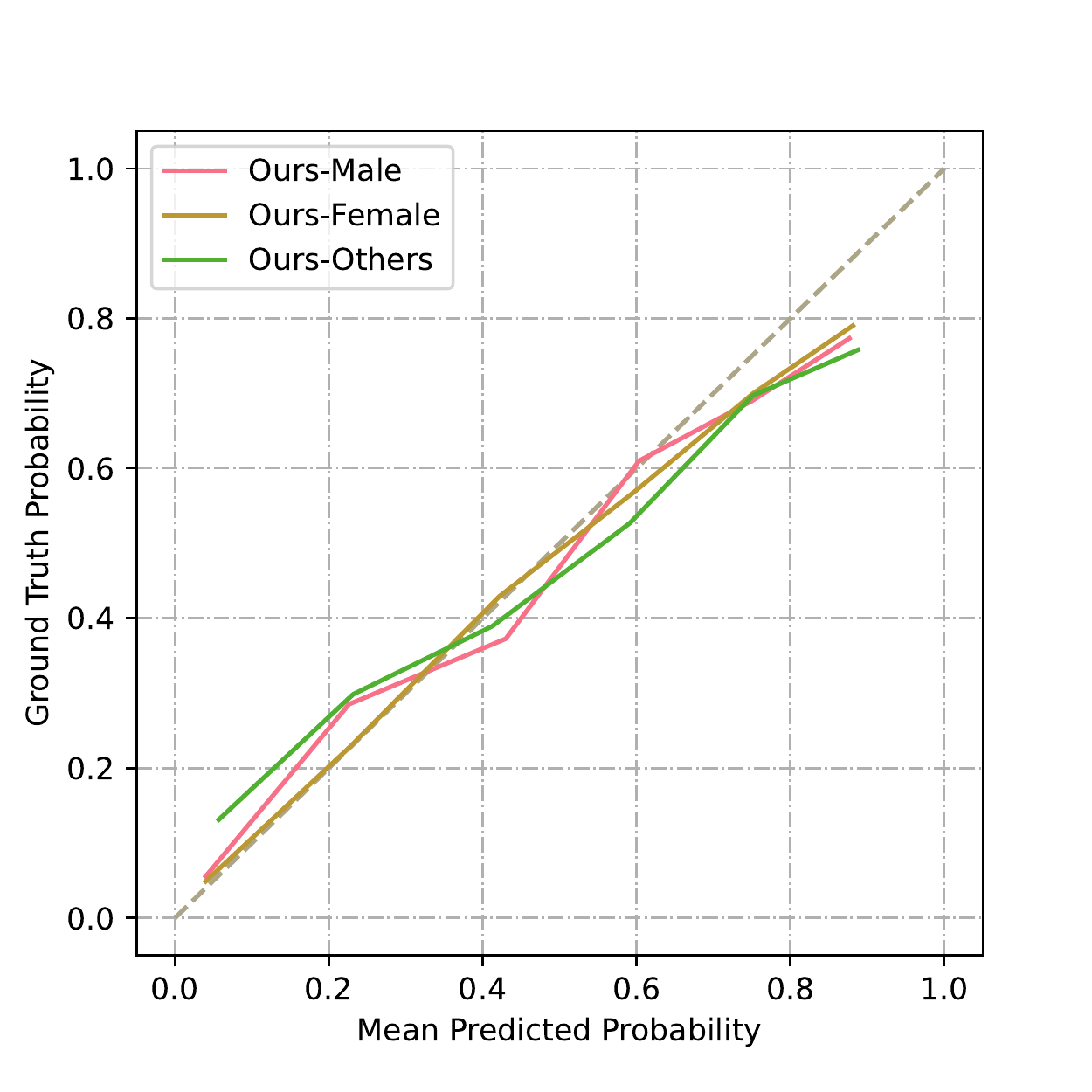}}
  \caption{Ablation. Results in Toxic dataset (Gender). Visualization of probability calibration for each subgroup $A=a$, i.e. $(f(X), \E[Y|f(X),A=a])$. The proposed approach shows significant improved probability calibration for each subgroup.}\label{fig:toxic_gender_proba_cal}
\end{figure}

The additional results of each subgroup's fair performance and the probability calibration on the Toxic Comments (Gender) dataset are shown in Fig.~\ref{fig:toxic_gender} and Fig.~\ref{fig:toxic_gender_proba_cal}.

It is worth noting that although the group sufficiency in three approaches is quite similar. However, the proposed approach shows a significant better probability calibration than baselines. 

\subsection{Results on different subgroup numbers}
\begin{figure}[h]
\centering
  \subfigure[Subgroup Number: 50]{\includegraphics[width=45mm]{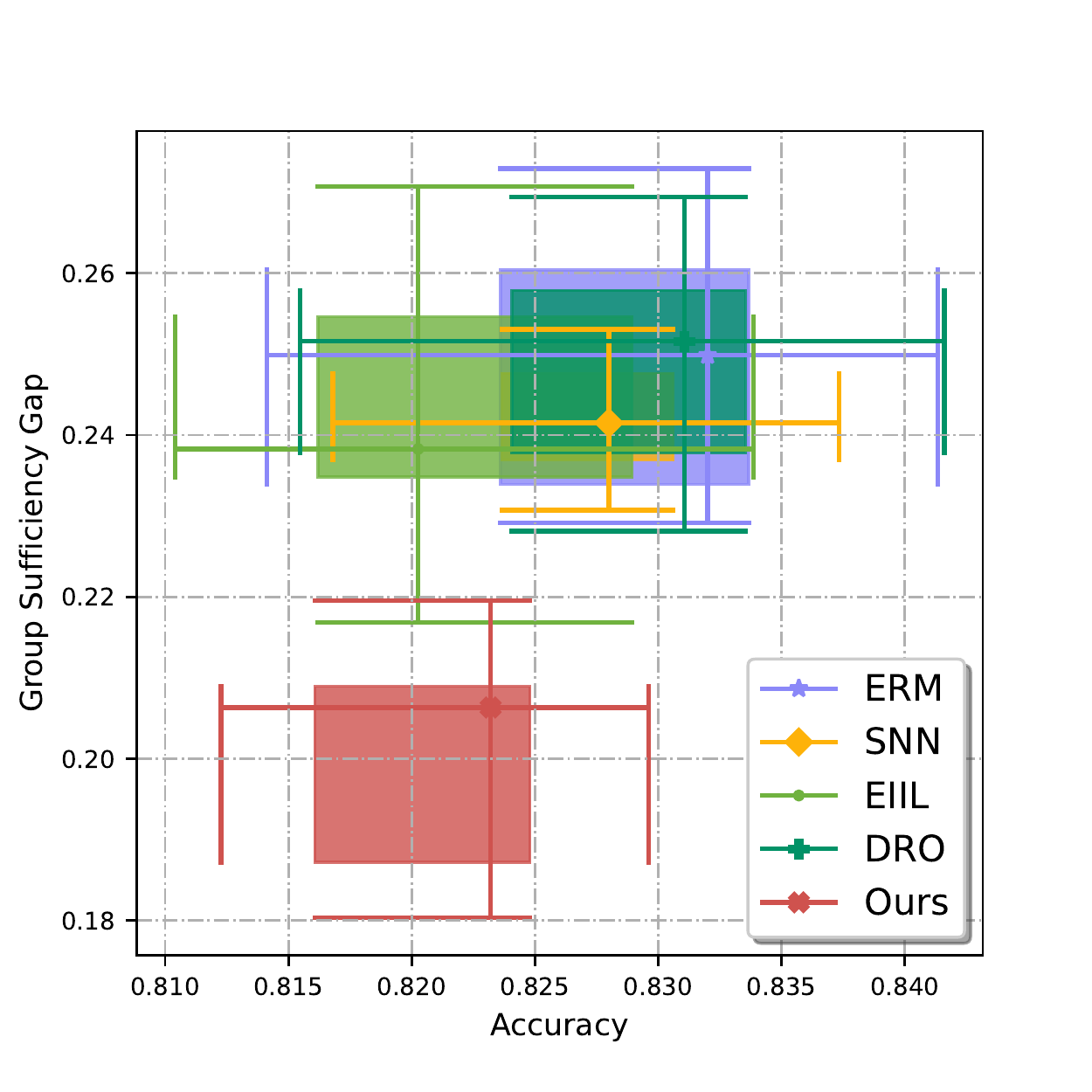}}
  \quad
  \subfigure[Subgroup Number: 100]{\includegraphics[width=45mm]{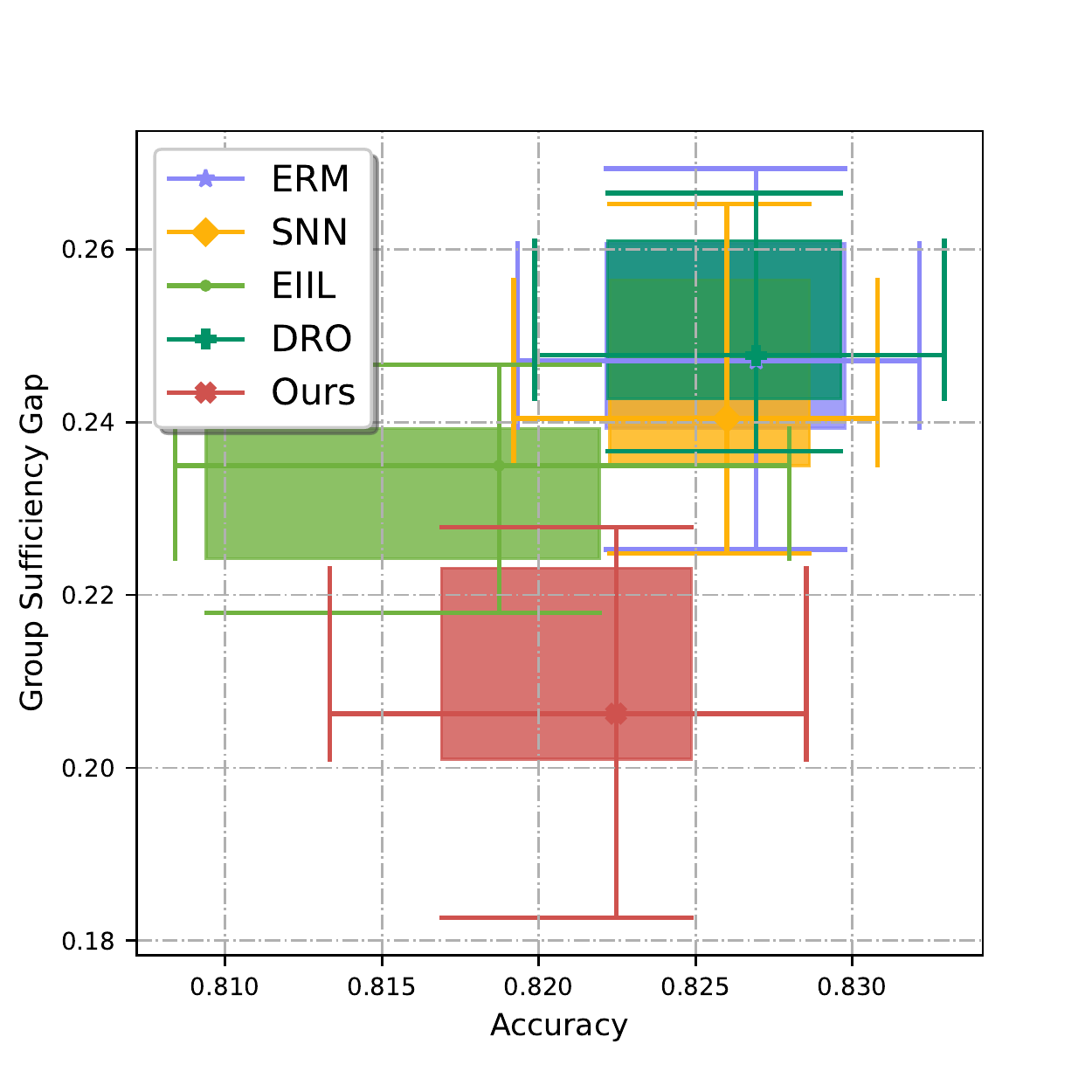}}
  \quad
  \subfigure[Subgroup Number: 300]{\includegraphics[width=45mm]{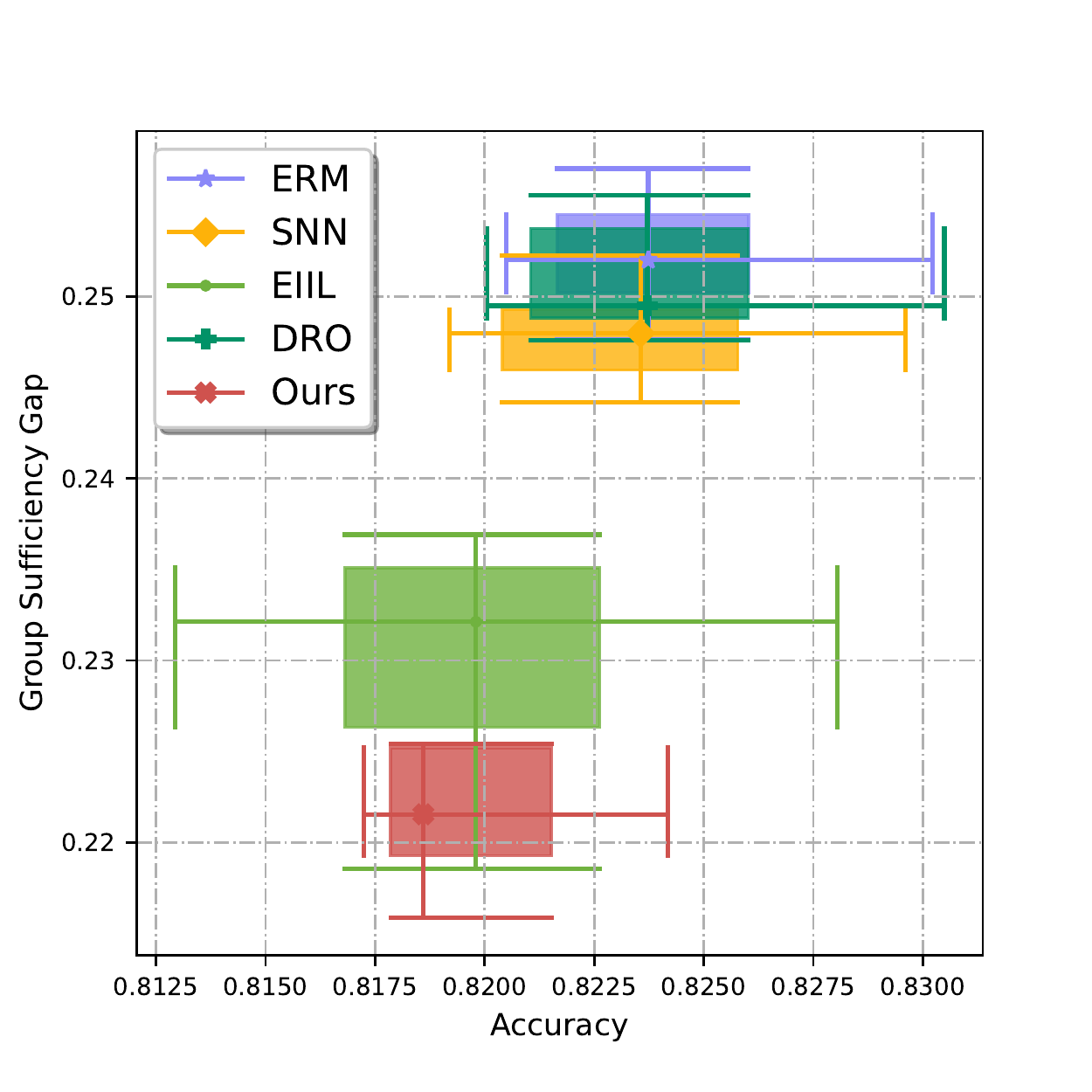}}
  \quad
  \subfigure[Subgroup Number: 400]{\includegraphics[width=45mm]{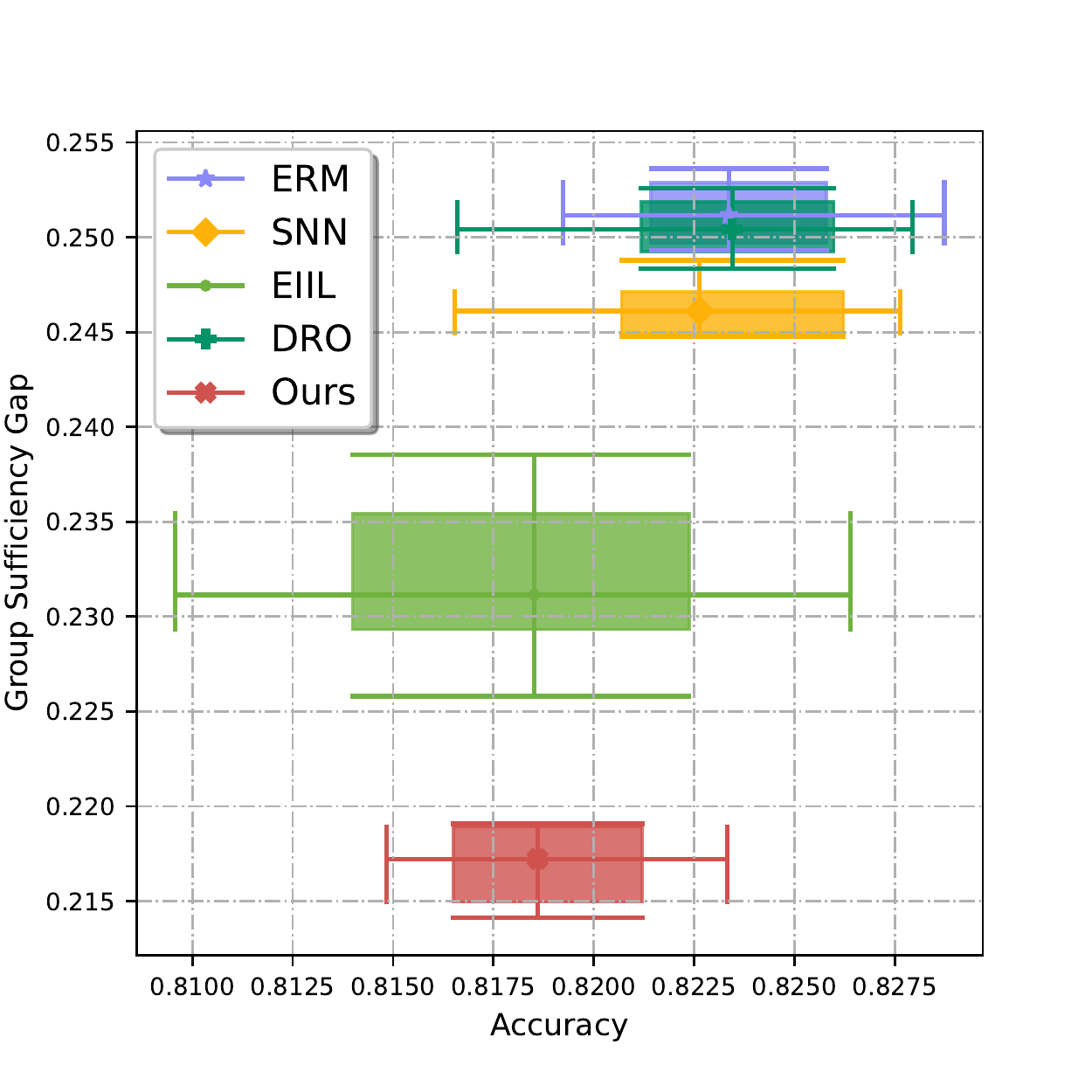}}
  \quad
  \caption{Results of different subgroups numbers on Amazon Review Dataset }\label{fig:amazon_diff_subgroup_numbers}
\end{figure}

We visualize the results on different subgroup numbers, shown in Fig.~\ref{fig:amazon_diff_subgroup_numbers}. The results still suggest the consistently better results than baselines. 

\subsection{Additional Results on Adult dataset}
We further evaluated the Adult dataset \cite{Dua:2019}, which predicts salary being larger than 50K or not. We treat \emph{gender} as the sensitive attribute and subsample 500 samples for each subgroup. We adopt  $\tilde{f}_{\bf{w}}$ and $\tilde{f}_{\bf{w}_a}$ as the two-layer fully connected neural network, where $ {\bf w}\sim Q$ and  $ {\bf w}_a\sim Q_a$. All results are repeated 4 times and illustrated in Tab.~\ref{tal:res_adult}.

\begin{table}[h]
\caption{Results (average and std) in Adult dataset (on \%)}
\centering
\begin{tabular}{@{}l|c|c@{}}
\toprule
Method & Accuracy  & Group sufficiency gap (smaller is better)  \\ \midrule
ERM & 83.38 (0.421) & 1.096 (0.182)   \\ \midrule
SNN & 82.62 (0.478) & 1.071 (0.220)   \\ \midrule
EIIL& 83.27 (0.547) & 1.018 (0.239)   \\ \midrule
FSCS& 83.63 (0.704) & 1.271 (0.316)   \\ \midrule
DRO & 83.48 (0.377) & 1.056 (0.151)   \\ \midrule
Ours & 83.33 (0.308) &  0.684 (0.008)  \\ \bottomrule
\end{tabular}
\label{tal:res_adult}
\end{table}
The result suggests a consistently better group sufficiency with comparable accuracy.

\subsection{Additional Results on vision dataset}
We further consider CelebA dataset as a computer vision task \cite{liu2015faceattributes}. We follow the protocol of \cite{chuang2021fair}, which predicts the wavy hair $Y$ in the image $X$.
We regard gender as sensitive attribute $A$. We further adopted Res18 as the backbone and three layers fully-connected (randomized) layers. We sub-sample 200 instances per subgroup and fine tune for maximum 20 epochs. All results are repeated 4 times and illustrated in Tab.~\ref{tal:res_celeba}.

\begin{table}[h]
\caption{Results (average and std) in CelebA dataset (on \%)}
\centering
\begin{tabular}{@{}l|c|c@{}}
\toprule
Method & Accuracy  & Group sufficiency gap (smaller is better)  \\ \midrule
ERM & 79.67 (0.40)  & 7.13 (0.96)   \\ \midrule
SNN & 79.79 (0.33) & 7.12 (0.89)   \\ \midrule
EIIL& 79.90 (0.30) & 6.23 (1.52)   \\ \midrule
FSCS& 79.33 (0.62) & 7.01 (1.88)   \\ \midrule
DRO & 79.82 (0.27) & 6.68 (1.14)   \\ \midrule
Ours & 79.87 (0.17) &  5.29 (0.67)  \\ \bottomrule
\end{tabular}
\label{tal:res_celeba}
\end{table}
The result also suggests a consistently better group sufficiency with comparable accuracy.

\subsection{Evolution of Q during the training}
We visualize the test accuracy and group sufficiency gap of fair predictor $Q$ during the training, shown in Tab.~\ref{tal:training_dynamic}. The results are evaluated on the Toxic data with race as the sensitive attribute. 

\begin{table}[h]
\caption{Evolution of accuracy and group sufficiency gap during the training (on \%)}
\centering
\begin{tabular}{@{}l|c|c|c|c|c|c|c|c@{}}
\toprule
Epoch & 0  & 2 & 4 & 6 & 8 & 10 & 12 & 14  \\ \midrule
Suf gap  & 12.49  &  14.65 & 11.60 & 7.32 & 6.19 & 5.41 & 5.45 & 5.42 \\ \midrule
Accuracy & 50.94  &  66.30 & 87.87 & 91.71 & 92.10 & 92.10 & 92.16 & 92.13 \\ \bottomrule
\end{tabular}
\label{tal:training_dynamic}
\end{table}

\end{document}